%% file: main.tex
\documentclass[10pt,twocolumn,letterpaper]{article}

\usepackage[pagenumbers]{cvpr}

\input{preamble}

\definecolor{cvprblue}{rgb}{0.21,0.49,0.74}
\usepackage[pagebackref,breaklinks,colorlinks,allcolors=cvprblue]{hyperref}

\usepackage{xurl}
\hypersetup{
  colorlinks,
  linkcolor = BrickRed,
  citecolor = RoyalBlue,
  urlcolor  = WildStrawberry,
}

\AddToHook{cmd/appendix/before}{%
    \crefalias{section}{appendix}%
    \crefalias{subsection}{appendix}
}

\title{Suppressing Non-Semantic Noise in Masked Image Modeling Representations}

\author{Martine Hjelkrem-Tan$^{1}$,
Marius Aasan$^{1}$, Rwiddhi Chakraborty$^{2}$,\\ Gabriel Y. Arteaga$^{1}$, Changkyu Choi$^{1}$, Ad\'in Ram\'irez~Rivera$^{1}$ \\
{\small
$^{1}$University of Oslo, 
$^{2}$UiT The Arctic University of Norway
} \\
{\scriptsize \texttt{
\{matan, mariuaas, gabrieya, changkyc, adinr\}@uio.no, rwiddhi.chakraborty@uit.no
}}
}

\begin{document}

\input{sec/main_content}

{
    \small
    \bibliographystyle{ieeenat_fullname}
    \bibliography{abrv, main}
}

\include{sec/X_suppl}

\end{document}

%% file: preamble.tex
\newcommand{\TODO}[1]{\textbf{\color{red}[TODO: #1]}}
\renewcommand{\TODO}[1]{}

\usepackage{fontawesome6}
\usepackage[T1]{fontenc}
\usepackage[utf8]{inputenc} 

\usepackage{tabularray}
\UseTblrLibrary{
  booktabs,
}

\usepackage[]{caption} 
\usepackage[dvipsnames, table]{xcolor}
\usepackage{graphicx}
\usepackage{wrapfig}
\usepackage{adjustbox}
\usepackage{tikz}
\usepackage{pgfplots}
\usepackage{pgffor}
\usetikzlibrary{
    arrows.meta,
    calc,
    fit,
    matrix,
    positioning,
    shadows,
    tikzmark,
}
\usepgfplotslibrary{
    groupplots,
    colorbrewer,
}
\pgfplotsset{
  compat=1.18,
  cycle list/Dark2,
}

\pgfdeclarelayer{background}
\pgfdeclarelayer{foreground}
\pgfsetlayers{background,main,foreground}

\colorlet{highlight}{BurntOrange!15}

\makeatletter
\patchcmd{\NAT@test}{\else \NAT@nm}{\else \NAT@nmfmt{\NAT@nm}}{}{}

\DeclareRobustCommand\citepos
  {\begingroup
   \let\NAT@nmfmt\NAT@posfmt
   \NAT@swafalse\let\NAT@ctype\z@\NAT@partrue
   \@ifstar{\NAT@fulltrue\NAT@citetp}{\NAT@fullfalse\NAT@citetp}}

\let\NAT@orig@nmfmt\NAT@nmfmt
\def\NAT@posfmt#1{\NAT@orig@nmfmt{#1's}}
\makeatother

\makeatletter
\def\NAT@spacechar{~}
\makeatother

\usepackage{xspace}
\makeatletter
\DeclareRobustCommand\onedot{\futurelet\@let@token\@onedot}
\def\@onedot{\ifx\@let@token.\else.\null\fi\xspace}

\def\cf{{cf}\onedot}

\makeatother

\makeatletter
\DeclareRobustCommand{\METHODNAME}{SOAP\texorpdfstring{\@ifnextchar.{\@}{\xspace}}{}}
\DeclareRobustCommand{\METHODFULLNAME}{Semantically Orthogonal Artifact Projection\xspace}
\makeatother

\usepackage{mathtools}
\usepackage{amsmath, amsfonts, bm}

\newcommand{\1}{\bm{1}}
\newcommand{\R}{\mathbb{R}}
\newcommand{\Rho}{\mathrm{P}}
\newcommand{\score}{\ensuremath{\text{SI}}}

\let\given\givenbase

%% file: sec/main_content.tex
\maketitle

\begin{abstract}
Masked Image Modeling (MIM) has become a ubiquitous self-supervised vision paradigm. In this work, we show that MIM objectives cause the learned representations to retain non-semantic information, which ultimately hurts performance during inference. We introduce a model-agnostic score for semantic invariance using Principal Component Analysis (PCA) on real and synthetic non-semantic images. Based on this score, we propose a simple method, \METHODFULLNAME (\METHODNAME), to directly suppress non-semantic information in patch representations, leading to consistent improvements in zero-shot performance across various MIM-based models. \METHODNAME is a post-hoc suppression method, requires zero training, and can be attached to any model as a single linear head.
Code available at: {\small \url{https://github.com/dsb-ifi/soap}}.
\end{abstract}

\section{Introduction}
\label{sec:introduction}

Self-supervised learning (SSL) via Masked Image Modeling (MIM) objectives have become a popular source for strong, generalized vision backbones~\citep{zhou2022ibot,caron2021emerging, oquab2024dinov2, simeoni2025dinov3,darcet2025cluster,tschannen2025siglip2,vekataramanan2025franca,assran2023ijepa,he2022mae,data2vec,mimrefiner}.
However, recent works have uncovered key issues with artifacts and noise in the representations in models that rely on MIM-based objectives~\citep{vekataramanan2025franca, zhou2022ibot, oquab2024dinov2, simeoni2025dinov3, darcet2025cluster, przewiezlikowski2025beyond}.
Some of these issues can be traced to the objective itself---MIM requires predicting both the semantic content and location of the masked patches~\citep{bar2024stop}.
While positional collapse---where the model learns to predict the position of masked tokens instead of content---is a known issue with various suggested mitigation techniques~\citep{darcet2025cluster, bar2024stop, vekataramanan2025franca}, studies directly addressing the extent of this phenomenon are relatively unexplored.

\begin{figure*}[tb]
\centering
\colorlet{str}{Dark2-A}
\colorlet{sem}{Dark2-B}
\colorlet{win}{BurntOrange}
\begin{tikzpicture}[
  node distance=.5cm,
  every node/.append style={font=\small},
  bag/.style={
    matrix of nodes,
    nodes in empty cells,
    nodes={
      pc,
      sharp corners,
    },
    row sep=3pt,
    column sep=3pt,
    draw,
    rounded corners,
  },
  pc/.style={
    draw,
    thick,
    minimum size=.5cm,
    text width=.28cm,
    inner sep=0pt,
    outer sep=0pt,
    align=center,
    font=\footnotesize,
  },
  del/.style={
    draw opacity=0.25,
    fill opacity=0.25,
  },
  proc/.style={
    draw=#1,
    fill=#1!50,
    minimum width=1.8cm,
    text width=1.8cm,
    align=center,
    minimum height=.8cm,
    rounded corners,
  },
  edg/.style={
    ->,
    shorten <= 2pt,
    shorten >= 2pt,
    rounded corners,
  },
  lbl/.style={
    black!75,
    outer sep=0pt,
  },
  img frm/.style 2 args={
    rounded corners=2pt,
    clip,
    inner sep=0pt,
    outer sep=0pt,
    append after command={
        node [draw=#1, very thick, fit=(\tikzlastnode), inner sep=1pt, rounded corners=2pt] (\tikzlastnode-frm) {}
        node [#1, draw=#1, fill=white, very thick, circle, font=\small, inner sep=1pt] at (\tikzlastnode.north east) {#2}
    },
  },
  soap/.style={
    draw=#1,
    fill=#1!50,
    minimum width=1.5cm,
    text width=1.5cm,
    align=center,
    minimum height=.8cm,
    rounded corners,
    append after command={
      node[fill=white, draw=white, circle, minimum size=14pt, inner sep=0pt] (\tikzlastnode-) at (\tikzlastnode.35) {}
      node[fill=#1!50, draw=none, circle, minimum size=10pt, inner sep=0pt] (\tikzlastnode-1) at (\tikzlastnode) {}
      node[fill=#1!50, above left=2.5pt and 5pt of \tikzlastnode, circle, minimum size=8pt, inner sep=0pt](\tikzlastnode2) {}
      node[fill=#1!50, above right=-1pt and 6pt of \tikzlastnode, circle, minimum size=8pt, inner sep=0pt](\tikzlastnode3) {}
    },
  },
  tag/.style={
      font=\footnotesize,
      inner sep=1pt,
      outer sep=1pt,
  },
  proc tag/.style={
      tag,
      fill=white,
      rounded corners,
      fill opacity=0.75,
      text opacity=1,
  },
  patch/.style={
        draw,
        minimum width=0.3cm,
        minimum height=0.3cm
  },
  /cuboid/width/.store in=\cubewidth,
  /cuboid/height/.store in=\cubeheight,
  /cuboid/depth/.store in=\cubedepth,
  /cuboid/color/.store in=\cubecolor,
  cuboid/.pic={
    \tikzset{/cuboid/.cd,
      width=.15,
      height=.15,
      depth=.75,
      color=Dark2-D,
      #1}
    \colorlet{frontface}{\cubecolor!80}
    \colorlet{rightface}{\cubecolor!60}
    \colorlet{topface}{\cubecolor!40}
    \colorlet{edgecolor}{\cubecolor!40!black}
    \coordinate (-a) at (0,0,0);
    \coordinate (-b) at (\cubewidth,0,0);
    \coordinate (-c) at (\cubewidth,\cubeheight,0);
    \coordinate (-d) at (0,\cubeheight,0);
    \coordinate (-e) at (\cubewidth,0,-\cubedepth);
    \coordinate (-f) at (\cubewidth,\cubeheight,-\cubedepth);
    \coordinate (-g) at (0,\cubeheight,-\cubedepth);
    %
    \path[fill=rightface] (\cubewidth,0,0) -- ++(0,0,-\cubedepth) -- ++(0,\cubeheight,0) -- ++(0,0,\cubedepth) -- cycle;
    %
    \path[fill=topface] (0,\cubeheight,0) -- ++(\cubewidth,0,0) -- ++(0,0,-\cubedepth) -- ++(-\cubewidth,0,0) -- cycle;
    %
    \path[fill=frontface] (0,0,0) -- ++(\cubewidth,0,0) -- ++(0,\cubeheight,0) -- ++(-\cubewidth,0,0) -- cycle;
    \path[draw=edgecolor, thin, shorten <=.5\pgflinewidth,shorten >=.5\pgflinewidth, miter limit=2] (0,\cubeheight,0) -- ++(\cubewidth,0,0) -- ++(0,-\cubeheight,0) -- ++(-\cubewidth,0,0) -- ++(0,\cubeheight,0) -- ++(0,0,-\cubedepth) -- ++(\cubewidth,0,0) -- ++(0,-\cubeheight,0) -- ++(0,0,\cubedepth) -- ++(0,\cubeheight,0) -- ++(0,0,-\cubedepth);
  },
  /matrix cuboid/x step/.store in=\xstep,
  /matrix cuboid/y step/.store in=\ystep,
  cuboid matrix/.pic={%
    \tikzset{/matrix cuboid/.cd,
      x step=0.2,
      y step=0.2,
      cuboids opts/.style={},
      #1}
      \foreach \i in {0,1,2,3} {
        \foreach \j in {3,2,1,0} {
          \pic (-\i-\j) at (\i*\xstep,-\j*\ystep) {cuboid={/matrix cuboid/cuboids opts}};
        }
      }
      \path let
        \p1 = (-0-1-d),
        \p2 = (-3-1-c),
        \p3 = (-0-1-a),
        \p4 = (-0-2-d),      
        \p5 = ($(\p1)!0.5!(\p2)$),
        \p6 = ($(\p3)!0.5!(\p4)$),
        \p7 = (\p6 -| \p5)
      in
        coordinate (-center) at (\p7);
    %
    \node[fit=(-0-0-d)(-3-3-b), inner sep=0] (-frm) {};
    \node[fit=(-0-0-g)(-3-3-e), inner sep=0] (-3dfrm) {};
  },
]
    \node[] (i) {\includegraphics[width=3cm]{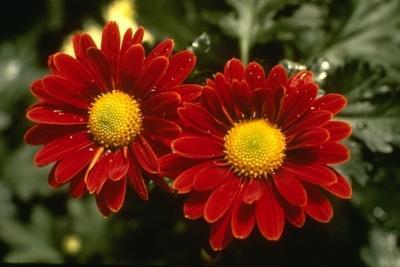}};

    \node[right=1cm of i, proc=Dark2-C] (enc) {MIM\\Encoder};

    \draw[pic anchor=(-center)] pic (z-p) at ($(enc.east)+(2cm,0)$) {cuboid matrix};
    \pgfnodealias{z}{z-p-frm}
    
    \node[lbl, above=-4pt of z, xshift=-5mm] (zl) {\normalsize{$z$}};

    \node[below=0.7cm of z, fill=Dark2-E!50, inner sep=6.5pt, soap=Dark2-E] (mth) {\METHODNAME};

    \draw[pic anchor=(-center |- -0-0-g)] pic (zc-p) at ($(mth.south)+(0,-.5cm)$) {cuboid matrix={cuboids opts/.append style={color={Dark2-A}}}};
    \pgfnodealias{zc}{zc-p-frm}

    \node[at=(zc-p-frm.north east)] (zcf) {
        \resizebox{1cm}{1cm}{%
        \includegraphics{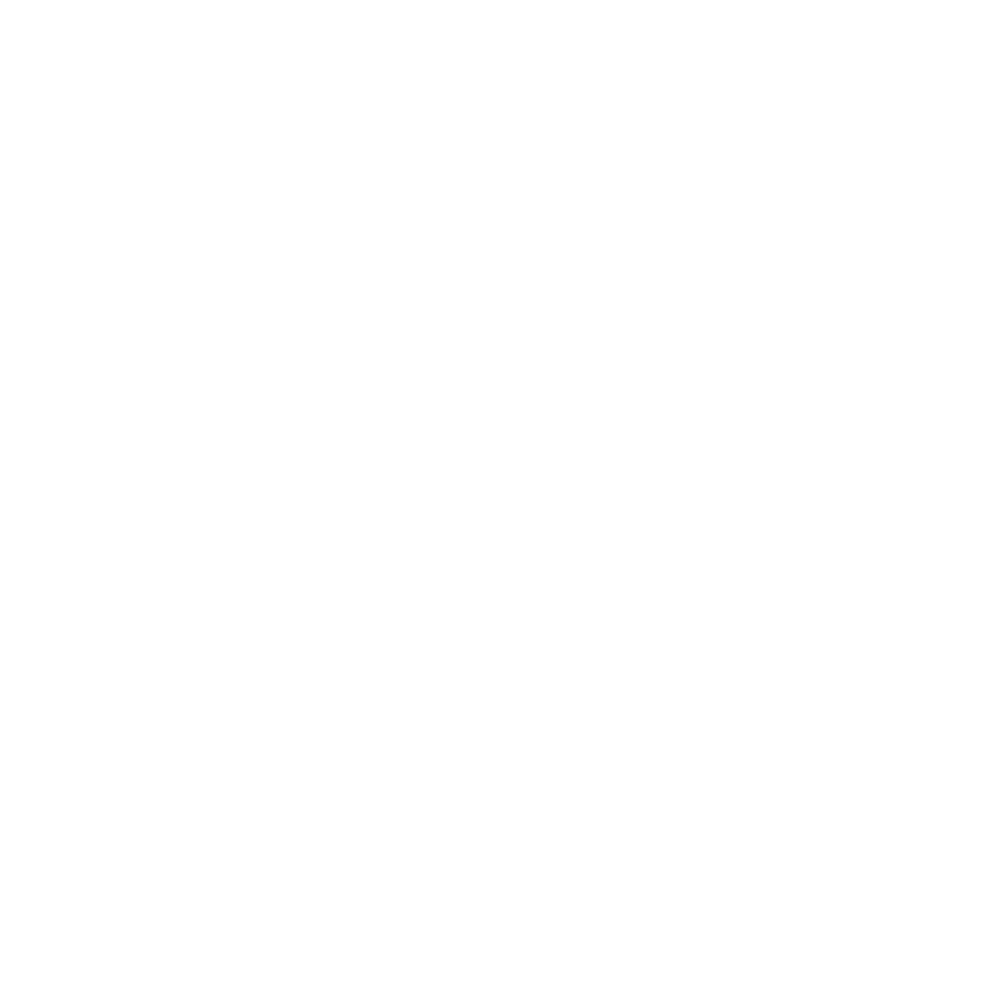}
        }
    };

    \node[lbl, above=-4pt of zc, xshift=-5mm] (zcl) {\normalsize{$\hat z$}};

    \node[right=2cm of z, img frm={Red}{\faIcon[regular]{thumbs-down}}] (o-o){\includegraphics[width=3cm]{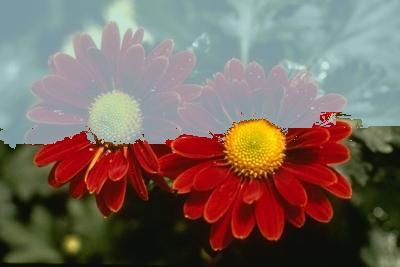}};
    \node[right=2cm of zc, img frm={Green}{\faIcon[regular]{thumbs-up}}] (o-a) {\includegraphics[width=3cm]{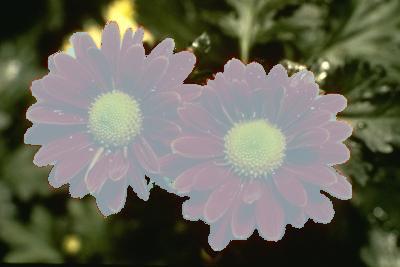}};

    \node[proc tag, inner sep=2pt, fill=Aquamarine!25, xshift=0.5cm] (hlbl) at ($(o-o)!.5!(o-a)$) {\small{Salient Prediction}};
    \draw[edg, -{Circle[length=2pt]}, copy shadow={draw=white, shadow xshift=.5pt, shadow yshift=-.5pt}] (hlbl) -- ($(o-o.north east)+(225:10pt)$);
    \draw[edg, -{Circle[length=2pt]}, copy shadow={draw=white, shadow xshift=.5pt, shadow yshift=-.5pt}] (hlbl) -- ($(o-a.center)$);

    \node[
    at=(i.south west |- o-a.south),
    shift={(8pt,8pt)},
    anchor=south west,
    align=left,
    text width=2.5cm,
    font=\footnotesize,
    ] (b-text) {%
      Principal components\\
      with \textcolor{str}{positional noise}\\
      and \textcolor{sem}{semantic info}. \\
      \METHODNAME removes noise.
    };
    \matrix (b-o) [right= of b-text, bag] {%
      |[str]|\faIcon{arrow-left} & |[str]|\faIcon{arrow-down} & |[str]|\faIcon{arrow-right} \\ 
      |[str]|\faIcon{arrows-split-up-and-left} & |[sem]|\faIcon{border-all} & |[sem]|\faIcon{circle} \\
      |[sem]|\faIcon{circle-half-stroke} & |[sem]|\faIcon{circle-nodes} &|[sem]|\faIcon{cookie} \\
    };
    \node[
      draw=Dark2-E!80,
      dashed,
      very thick,
      rounded corners, 
      inner sep=4pt,
      fit=(b-text) (b-o)
    ] (sq) {};
    
    \draw[draw=Dark2-E!80, dashed, very thick, rounded corners] (sq.north east) |- (mth);
        
    \draw[edg] (i) -- (enc);
    \draw[edg] (enc) -- (z);
    \draw[edg] (z) -- (mth);
    \draw[edg] (mth) -- (zc |- zc-p-3dfrm.north);
    \draw[edg] (z.east -| z-p-3dfrm.east) -- (o-o) node[tag, pos=0.5, above] {\scriptsize{Salient seg.}};
    \draw[edg] (zc.east -| zc-p-3dfrm.east) -- (o-a) node[tag, pos=0.5, above] {\scriptsize{Salient seg.}};

    \node[lbl, above=1mm of o-o] (dst) {Downstream task};
    \node[lbl, above=5mm of z] (rep) {Repr.};
    \node[lbl, right=1mm of mth, gray] (nice) {\footnotesize\textit{Training Free!}}; 
\end{tikzpicture}

\caption{
Pipeline overview; a pretrained MIM encoder outputs dense representations $z$ which are used for downstream tasks---we show salient segmentation as an example.
By identifying and suppressing principal components encoding positional noise, our \METHODNAME module improves the representations $\hat z$ and enhances downstream performance in zero-shot settings.
}
\label{fig:diagram}
\end{figure*}
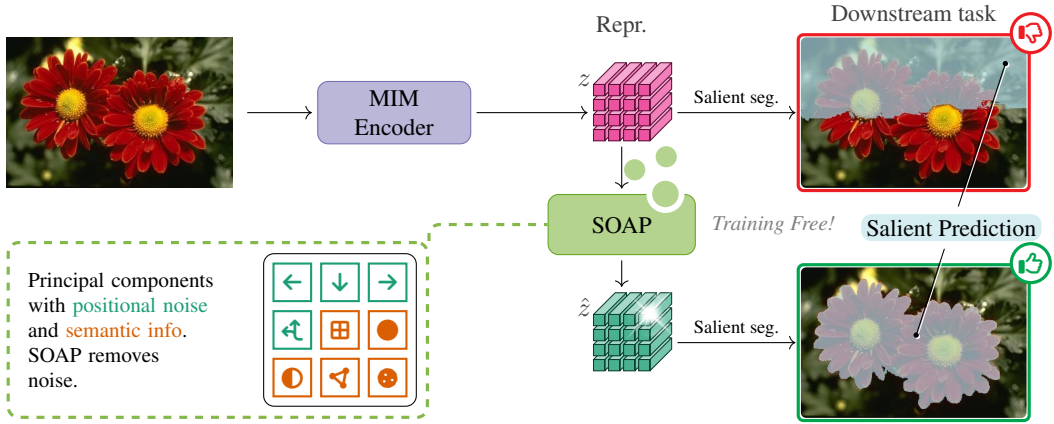

In this paper, we present a novel method to measure the amount of \emph{non-semantic noise} in ViT tokens for state-of-the-art MIM-based models. 
We characterize non-semantic noise as components that are invariant to the \emph{semantic content} in the input. 
This can for example be positional encoding, which are necessary for attention mechanisms but seldom useful in inference, or structural artifacts as discussed in \citet{darcetregister2024}.
Our central hypothesis in this work is that \emph{this noise persists in non-semantic images}. As a result, isolating the noise can be approached as quantifying invariant responses between semantic and non-semantic images. 
This is useful, as suppressing the noise can lead to clearer semantic signals, improving representations for use in downstream tasks.
Through our analysis, we discover that strong principal components exhibit high levels of non-semantic noise, and that this feature is \textit{pervasive in MIM-based models} while \textit{nearly non-existent in other, non MIM-based SSL models}.
Importantly, this holds \textit{regardless of which positional embedding method is employed}~\citep{simeoni2025dinov3} and whether predictions are conducted in latent or input-space, suggesting that this is an \textit{implicit issue in MIM}\@.

To suppress non-semantic information, we introduce a \METHODFULLNAME (\METHODNAME) to remove unwanted artifacts that are not useful for inherently semantic tasks, such as instance level classification and salient segmentation---\cf \cref{fig:diagram}.
\METHODNAME is flexible:
It is computed directly from data using a Gram-Schmidt based projection, thus requiring no training, and can be attached as an external module to any pretrained SSL backbone.

Our \textbf{contributions} include:
\begin{enumerate*}[label={(\roman*)}]
    \item An in-depth analysis that shows that MIM objectives \textit{uniquely} bias representations toward encoding positional noise rather than semantic information.
    \item A novel  \textit{Semantic Invariance Score} to measure the level of semantic invariance in a model’s representations, allowing us to diagnose the semantic-positional noise trade-off in SSL representations.
    \item Finally, based on this score, we propose \METHODFULLNAME (\METHODNAME), a post-hoc denoising strategy that suppresses non-semantic noise, and show improved performance in zero-shot downstream tasks for all MIM-based models.
\end{enumerate*}

\section{Related Work}
\label{sec:related_work}

\paragraph{Contrastive Learning.}
Contrastive learning by encouraging representations to be invariant to minor augmentations is a central paradigm for SSL~\citep{chen2020simple, tian2020good, chen2020mocov2, wang2020understanding, zhang2022rethinking}.
Modern approaches utilize self-distillation by matching representations between a teacher and a student network, with periodic exponential moving average (EMA) updates~\citep{grill2020byol, caron2020swav, chen2021simsiam, caron2021emerging}.
These approaches typically rely on strong global cues that match local with global views via aggressive cropping without preserving relative spatial information.
However, a stronger reliance on global cues limits performance on dense prediction tasks, as finer-grained semantic relationships are not fully exploited~\citep{liu2024unsupervised, yuan2023densedino, wang2022self, puy2024pillars}.

\paragraph{Masked Image Modeling.}
A strong alternative to learning invariances through data augmentations, is through choosing reconstruction of masked patches as the pretext task~\citep{he2022mae, xie2022simmim, bao2022beit}.
As opposed to contrastive learning, MIM---popularized by MAE~\citep{he2022mae}---provides a robust and scalable approach by directly reconstructing the input data via a decoder, thus circumventing representational collapse to trivial solutions.
Reconstruction in the representation space, rather than the pixel space, removes the explicit need for a decoder, and has been shown to work in methods like I-JEPA~\citep{assran2023ijepa}.
The MIM objective has been adopted in the self-distillation setup quite successfully---iBOT~\citep{zhou2022ibot} employs the averaged cross entropy over the student's masked view, and the teacher the non-masked view, and DINOv2 demonstrates additional improvements by separating the MLP projection heads, using adaptive resolutions, and additional regularization~\citep{oquab2024dinov2}. 
CAPI~\citep{darcet2025cluster} adopts a pure MIM approach through clustering by generating pseudo-labels in the masked latent space, which removes the need for a delicate balancing act between MIM and instance based contrastive objectives. 
The recent release of DINOv3 employs modern positional encoding, and a denoised warmup training phase~\citep{simeoni2025dinov3}. 
Across all these approaches, our analysis indicates a consistent level of non-semantic noise not present in purely contrastive approaches.

\paragraph{Positional Noise in MIM Models.}
A recurring challenge in masked image modeling (MIM) frameworks is the presence of positional noise in the learned embeddings.
This has been indirectly addressed through a variety of architectural and training modifications, such as introducing register tokens~\citep{darcetregister2024}, regularization mechanisms~\citep{darcet2025cluster, simeoni2025dinov3}, or disentangling positional cues~\citep{vekataramanan2025franca, yang2024denoising, bar2024stop, wang2024sinder}.
Despite these efforts, MIM representations typically show weaker out-of-the-box performance compared to frameworks with contrastive objectives, and often require extensive fine-tuning to transfer effectively to downstream tasks. 
Some works analyze embedding variance to identify noisy or unstable tokens~\citep{vanyan2024}, while others propose selective aggregation to prune non-informative dimensions~\citep{przewiezlikowski2025beyond}.
\citet{vekataramanan2025franca} introduce RASA, a post-hoc module that suppresses explicit location cues by training to predict patch positions from pretrained embeddings.
However, RASA only addresses patch-location noise and does not provide a broader analysis of non-semantic noise in other frameworks. 
In contrast, our proposed solution requires no training. Additionally, our study directly identifies non-semantic components rather than learning positional cues, probing the nature of semantically invariant noise encoded by state-of-the-art SSL models, and proposing metrics to quantify and interpret these effects.

\begin{figure*}[t]
    \centering
    \setlength{\tabcolsep}{2pt}
    \begin{tabular}{rcccccc@{\hskip 6pt}@{\hskip 6pt}cc}
        &
        \shortstack{\small \textcolor{Dark2-B}{DINOv2} \\ \footnotesize PC1, PC3} & 
        \shortstack{\small \textcolor{Dark2-B}{DINOv3} \\ \footnotesize PC2, PC3} & 
        \shortstack{\small \textcolor{Dark2-B}{iBOT} \\ \footnotesize PC2, PC4} & 
        \shortstack{\small \textcolor{Dark2-B}{MAE} \\ \footnotesize PC2, PC3} & 
        \shortstack{\small \textcolor{Dark2-B}{CAPI} \\ \footnotesize PC3, PC11} & 
        \shortstack{\small \textcolor{Dark2-B}{I-JEPA} \\ \footnotesize PC2, PC4} & 
        \shortstack{\small \textcolor{Dark2-A}{DINO} \\ \footnotesize PC1, PC2} & 
        \shortstack{\small \textcolor{Dark2-A}{DeiT3} \\ \footnotesize PC1, PC2} \\ 
        
        \rotatebox[origin=l]{90}{\footnotesize \text{Left / Right}} &
        \includegraphics[width=0.1\linewidth]{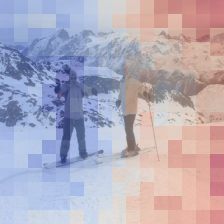} &
        \includegraphics[width=0.1\linewidth]{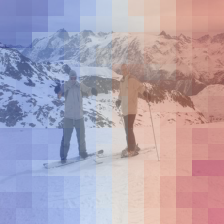} &
        \includegraphics[width=0.1\linewidth]{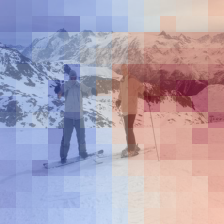} &
        \includegraphics[width=0.1\linewidth]{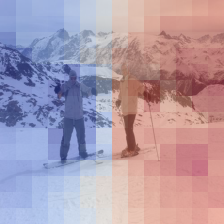} & 
        \includegraphics[width=0.1\linewidth]{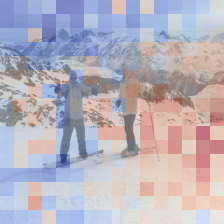} &
        \includegraphics[width=0.1\linewidth]{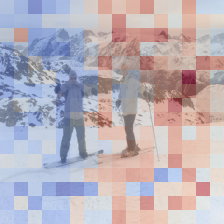} &
        \includegraphics[width=0.1\linewidth]{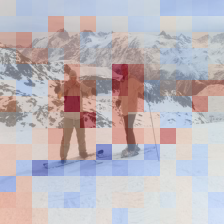} & 
        \includegraphics[width=0.1\linewidth]{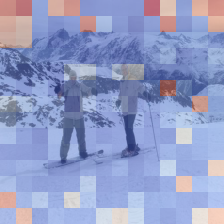} \\

        \rotatebox[origin=l]{90}{\footnotesize \text{Top / Bottom}} &
        \includegraphics[width=0.1\linewidth]{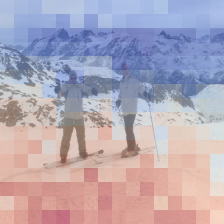} &
        \includegraphics[width=0.1\linewidth]{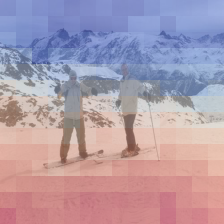} &
        \includegraphics[width=0.1\linewidth]{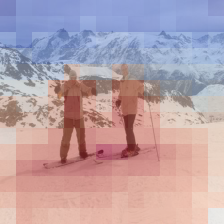} &
        \includegraphics[width=0.1\linewidth]{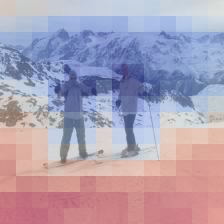} &
        \includegraphics[width=0.1\linewidth]{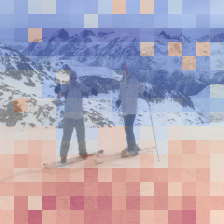} &
        \includegraphics[width=0.1\linewidth]{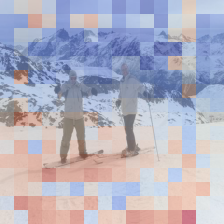} &
        \includegraphics[width=0.1\linewidth]{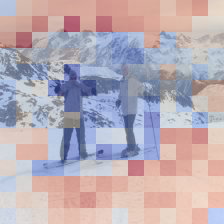} & 
        \includegraphics[width=0.1\linewidth]{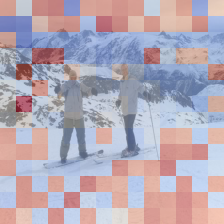} \\
    \end{tabular}
    \caption{
    Representations from MIM models exhibit strong positional bias in leading principal components (PC), illustrated here by the response heatmap of selected PCs for one example image. 
    \textcolor{Dark2-B}{MIM models} show clear left/right and top/bottom bias. This behavior is not observed for \textcolor{Dark2-A}{non-MIM models}. In Section \ref{sec:method}, we propose a novel method to automatically isolate such positional bias in a post-hoc fashion.
    }
    \label{fig:tblr_response}
\end{figure*}

\section{\METHODFULLNAME}
\label{sec:method}

Before introducing our proposal, we provide an exploratory analysis that sheds light on how to decompose the semantic and positional information from the data.  Specifically, we use principal component analysis (PCA) to reveal structural patterns that expose positional bias in MIM training (\cref{sec:pca}).     
Then, we formalize a linear decomposition of representations into semantic and non-semantic parts,  providing a foundation for quantifying the degree to which different components capture meaningful information versus noise (\cref{sec:lindecomp}).
Following, we propose a score to measure semantic invariance by contrasting real and synthetic inputs (\cref{sec:seminv}).
Finally, we introduce our \METHODFULLNAME (\METHODNAME), which uses this score to suppress non-semantic noise, enhancing zero-shot performance in MIM-based models on downstream tasks (\cref{sec:soap}).

\subsection{PCA of Patch Embeddings}
\label{sec:pca}
 
PCA is a natural tool to decompose the representation space into dominant variance directions. This allows us to examine whether leading components reflect semantic content or non-semantic structure, such as positional bias. 
To this end, we estimate the covariance of the $D$-dimensional patch embedding space $\mathcal{Z} \subseteq \mathbb{R}^D$ under a given model $f$ over ImageNet~\citep{imagenet} using Welford's online algorithm~\citep{welford1962note}, and obtain the principal components by eigendecomposition of the covariance matrix
\begin{equation}
    \text{Cov}(\mathcal{Z}) = V\Lambda V^\top \in \mathbb{R}^{D \times D},
\label{eq:pca}
\end{equation}
with principal component vectors $V=(v_1, \dots, v_D)$.

For each component $d=1,...,D$, we define the \emph{response} of an input~$z$ by the inner product $\langle z, v_d \rangle$, which reflects how strongly each patch embedding $z\in\mathcal{Z}$ projects onto direction $v_d$.
By inspecting the responses, we observe that leading principal components in MIM models exhibit strong biases based on patch position.
This indicates that a significant share of their variance is devoted to encoding position, \textit{implying that positional bias arises as a direct consequence of MIM objectives}.
\cref{fig:tblr_response} shows the responses for an example image for selected principal components that were identified as having a strong positional bias.
In contrast, models trained without MIM objectives, such as DINO and DeiT3, do not exhibit this behavior.

To further characterize these components, we evaluate their behavior on a set of images. 
Let $\Omega$ be the input set of images, and $f\colon\Omega \rightarrow \mathcal{Z}$ a model that encodes an image $x \in \Omega$ into $N$ patch embedding representations $\boldsymbol{z} = \{z_n\}_{n=1}^{N}$.
The binary activation of component $d$ for $x$ is then defined by thresholding the token responses:
\begin{align}\label{eq:activation}
    A_{d}(x)= \1 [ \boldsymbol{z} v_d  \geq \eta ] \in {\{0,1\}}^N.
\end{align}
Here $\eta$ is a scalar threshold; we set $\eta=0$ in our experiments.
The expectation $\bar{A}_d =\mathbb{E}_x[A_d(x)] \in \R^N$ yields an average activation map that highlights spatial patterns.

With these definitions in place, we formulate two diagnostic conditions to identify non-semantic components.
First, when the average activation $\bar{A}_d$ exhibits systematic dependence on patch location, we infer that component $d$ predominantly encodes positional cues rather than semantic content. 
Second, when the per-sample activations ${A}_{d}(x)$ exhibit minimal variation across different inputs, component $d$ can be regarded as invariant to image content, which indicates that it captures non-informative signals.

\subsection{Patch Embedding as Linear Combinations}
\label{sec:lindecomp}
To isolate the signals underlying a model’s representations, we express each patch embedding $z$ as a mixture of semantic and non-semantic sources.
Ideally, we want the model $f$ to capture the relevant semantic information in the input, like global class information and local structures of color, shape, texture, and fine-grained semantics useful for semantic downstream tasks.
However, in standard ViTs, $f$ must also encode the positional information directly to each embedding $z$, since the attention operator is otherwise permutation-invariant.

We formalize this by assuming that each embedding $z$ can be decomposed as a mixture of sources, $\Phi$ for semantic information and  $\Rho$ for non-semantic noise. 
Then we express
\begin{align}\label{eq:decomposision}
    z = \theta_\phi \phi + \underbrace{\theta_\rho \rho + \varepsilon}_{\mathclap{\text{non-semantic}}}; \quad \varepsilon \sim \mathbb{P},
\end{align}
where $\theta_\phi, \theta_\rho \in \mathbb{R}_{\geq 0}$ are scalar coefficients for $\phi \in \Phi, \rho \in \Rho$, and $\varepsilon$ represents residual noise from some probability distribution $\mathbb{P}$.  
The semantic information component $\phi\in\Phi$ encodes both relevant information for global objectives, such as instance discrimination and classification, and dense objectives like segmentation.
The non-semantic information component $\rho\in\Rho$ encodes local positions and relative geometry among the local patch embeddings $\boldsymbol{z}$, based on their positions in the original image $x$.\footnote{We note that positional information is given implicitly in certain models that retain spatial order, like CNNs and MLPs. In this case $f$ does not need to explicitly encode positional information into $z$. However, we restrict this study to ViTs.}

Intuitively, both contrastive and MIM-based training objectives encourage $f$ to encode relevant semantic information. However, in line with the observations of previous work~\citep{bar2024stop}, we posit that the inpainting objective in MIM also drives $f$ to encode a substantial amount of non-semantic information $\rho$, thereby increasing the coefficient $\theta_\rho$ at the expense of the semantic coefficient $\theta_\phi$. 
This raises the question of how much of a learned representation reflects semantic content, as opposed to non-semantic noise or other non-informative signals. 
To address this, we introduce a \emph{semantic invariance score} to measure the degree of non-semantic information encoded by each principal component $d$ in a learned patch representation space $\mathcal{Z}$.

\subsection{Semantic Invariance}
\label{sec:seminv}

Semantic invariance refers to the property of a component yielding consistent responses even when the semantic content of local representations varies.
In other words, a component is semantically invariant if it produces similar activations regardless of whether the input carries meaningful semantic information. We posit that such components are uninformative for downstream understanding, as they do not encode or reflect actual image contents.

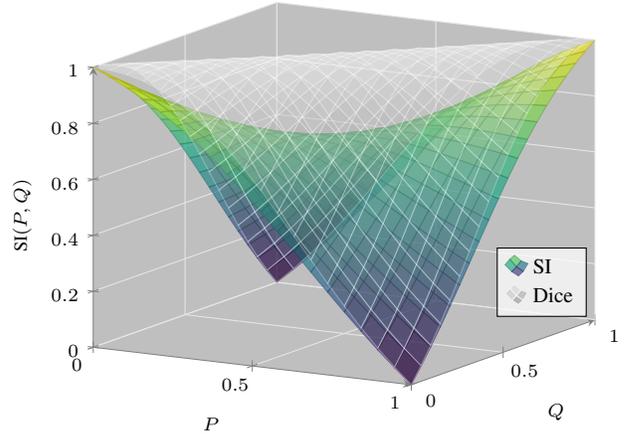
\begin{figure}[tb]
\centering
\begin{tikzpicture}
\begin{axis}[
    width=0.99\linewidth,
    height=0.8\linewidth,
    xlabel={$P$},
    ylabel={$Q$},
    zlabel={$\text{SI}(P,Q)$},
    xmin=0, xmax=1,
    ymin=0, ymax=1,
    zmin=0, zmax=1,
    xtick={0,0.5,1},
    ytick={0,0.5,1},
    view={30}{15},
    grid=major,
    grid style={gray!20},
    xlabel style={font=\scriptsize},
    ylabel style={font=\scriptsize},
    zlabel style={font=\scriptsize},
    tick label style={font=\scriptsize},
    axis lines=left,
    axis line style={gray},
    axis background/.style={fill=gray!50},
    legend style={
        cells={anchor=west},
        font=\footnotesize,
        at={(axis cs:.98,.99,0.01)},
        anchor=south east,
        fill opacity=0.75,
        text opacity=1,
    },
    every legend image post/.append style={
        scale=.5,
    }
]

\addplot3[
    surf,
    domain=0:1,
    domain y=0:1,
    samples=20,
    samples y=20,
    opacity=0.7,
    shader=faceted,
    colormap/viridis,
] 
{
2*(x * y + (1-x) * (1-y)) / ((x^2 + (1-x)^2)^(1/2) + (y^2 + (1-y)^2)^(1/2))
};
\addlegendentry{SI}

\addplot3[
    surf,
    domain=0:1,
    domain y=0:1,
    samples=20,
    samples y=20,
    opacity=0.3,
    shader=faceted,
    faceted color=white,
    colormap/blackwhite,
] 
{
2*(x * y + (1-x) * (1-y)) / (x^2 + (1-x)^2 + y^2 + (1-y)^2)
};
\addlegendentry{Dice}
\end{axis}
\end{tikzpicture}
\caption{
Plot of the semantic invariance (SI) score \textcolor{green!75!black}{(viridis)}.
The SI-score increases when $P\approx Q$ and the probabilities are confident (close to $0$ or $1$).
In the uncertain case $P\approx0.5\approx Q$, the score is lower to reflect ambiguity of semantic invariance.
For comparison we also show the Dice-Sørensen coefficient \textcolor{gray}{(gray)}, which does not have this property, and thus is unable to capture the uncertainty.
}
\label{fig:SI_score_3D}
\end{figure}

Let $\mathcal{X} \subset \Omega$ be the set of semantically informative images, and let $\mathcal{X}^{c}$ denote a complementary set without semantic information. 
In practice, $\mathcal{X}$ is instantiated as the ImageNet validation set~\citep{imagenet}, while $\mathcal{X}^{c}$ is approximated by a synthetic noise generator.
The synthetic images are generated by a mixture of pink noise, modulated white noise, and random low-frequency gradient fields. The details are described in \cref{sec:synth_data}---see the examples in \cref{fig:synth-data}.
For each component $d$, we compute binary activations $A_d$ using \cref{eq:activation} for samples $x \sim \mathcal{X}$ and $x^c \sim \mathcal{X}^{c}$.
This yields two empirical Bernoulli distributions of activations for the token index $n = 1,\dots,N$, such that
\begin{align}
P_{d,n} &= \text{Pr}\big( A_{d,n} = 1 \given x \sim \mathcal{X}\big), \quad \text{and}\\
Q_{d,n} &= \text{Pr}\big( A_{d,n} = 1 \given x^c \sim \mathcal{X}^c\big).
\end{align}
If $P_{d,n} \approx Q_{d,n}$, then component $d$ behaves similarly for semantic and non-semantic inputs for token index $n$, indicating semantic invariance.
Conversely, large discrepancies between $P_{d,n}$ and $Q_{d,n}$ indicate sensitivity to semantic content at the position with index $n$, which posits $P_d, Q_d$ as multinomial distributions over all tokens.
To quantify this discrepancy, we define a score to measure \textit{semantic invariance (SI)}
\begin{subequations}\label{eq:SIscore}
\begin{align}
    s_d &= \score(P_d,Q_d), \\ 
    &= 2 \cdot \frac{P_d \cdot Q_d + (1-P_d)\cdot(1-Q_d)}{\sqrt{P_d^2 + (1-P_d)^2} + \sqrt{Q_d^2 + (1-Q_d)^2}}, \\
    &= 2 \cdot \frac{ \langle {P}_d, {Q}_d \rangle}{||{P}_d|| + ||{Q}_d||},
\end{align}
\end{subequations}
which assigns high scores when $P_d \approx Q_d$, and vice versa.
However, in the case where $P_{d,n} \approx Q_{d,n} \approx 0.5$ for most token indices $n$, we cannot really say that the component is strongly semantically invariant, as the model is uncertain. 
We account for this uncertainty in our definition of the SI-score; it gives lower scores when the model is uncertain. 
This property is visualized in \cref{fig:SI_score_3D}.
Our score is similar in form to the Dice-Sørensen coefficient~\citep{carass2020evaluating}, but we take the inner products and norms over the support set rather than the multivariate dimensions; see \cref{sec:vector_si} for more details.
A vectorial formulation of \cref{eq:SIscore} is given in \cref{eq:geoSIscore}. 

\subsection{\METHODFULLNAME
(\METHODNAME)}
\label{sec:soap}
With the SI score in ~\cref{eq:SIscore}, we introduce \METHODFULLNAME
(\METHODNAME), an off-the-shelf denoising strategy that suppresses non-semantic noise.
We hypothesize that suppressing components that are invariant to semantic content in the representation $z$ acts as a denoising step, yielding representations that are closer to the semantic part $\theta_\phi \phi$ in \cref{eq:decomposision}.
To achieve this, we operate in the PCA basis, where these components $v_d$ are orthogonal by construction. 
Each component is assigned a weight $w_d$, derived from its semantic invariance score $s_d$ and a scaling function $t$, which determines how strongly it should be suppressed.
Our \METHODNAME projector $P_{\phi}$ is defined by the Gram–Schmidt process,
subtracting the contribution of components identified as non-informative
\begin{align}\label{eq:projection}
    P_{\phi} = I - VWV^\top; \quad \hat{z} = P_\phi z.
\end{align}
Here, $V$ are the PCA components from \cref{eq:pca}, $I$ is the identity matrix, and ${W=\text{diag}(w_1, \dots, w_D)}$ is a diagonal weight matrix.

Since directly weighting by the \score-scores leads to strong suppression across all components (see \cref{fig:SI_score_3D}) we introduce a scaling function $t$ so only the most invariant ones are suppressed.
We propose filtering using a Fermi window~\citep{fermiMRI1,fermiMRI2}---commonly used in MRI imaging---using the rank of the scores $r$. 
This corresponds to a smooth regularization of the scores~\citep{hansen} using a sigmoid gating approach~\citep{nguyen2024sigmoid} with explicit control over truncation and smoothness.
The scaling function is defined by
\begin{align}
    \label{eq:fermi}
    w_d = t(s_d, r) = s_d \times \frac{\sigma( (\mu - r) / \tau)}{\sigma(\mu/\tau)}; \hspace{.75em} r=\text{rank}(s_d).
\end{align}
Coupling statistical variance with semantic relevance, the hyperparameters $\mu$ and  $\tau$ provide flexible control of the cut-off and smoothness of suppression, providing a simple way to adapt denoising strength to the spectral structure of the embeddings.
{\cref{fig:weights}} shows the effect of the scaling function on the semantic invariance scores.

\section{Experiments}

For our study, we consider MAE~\citep{he2022mae}, I-JEPA~\citep{assran2023ijepa} and CAPI~\citep{darcet2025cluster} for models with MIM objectives, DINO~\citep{caron2021emerging} for global contrastive objective, and iBOT~\citep{zhou2022ibot}, DINOv2~\citep{oquab2024dinov2}, and DINOv3~\citep{simeoni2025dinov3} for models with both.
For completeness we also evaluate the supervised model DeiT3~\citep{deit3}.
\cref{tab:models} provides an overview of the models.

We estimate the principal components for the representation space of each model over ImageNet train~\citep{imagenet} as described in \cref{sec:pca}, and calculate the SI-scores using ImageNet validation paired with $50000$ synthetic images as described in \cref{sec:synth_data}.
\METHODNAME is directly computed from the principal components and corresponding SI scores.
We show strong improvements in zero-shot salient segmentation using the widely adopted TokenCut method~\citep{wang2023tokencut} when applying \METHODNAME prior to inference.
Further, we show improvements in kNN semantic segmentation and kNN classification using patch embeddings.
Additional details on our evaluation methods and ablations are given in \cref{sec:evaluation_details}.

\begin{table}[tb]
\centering
\footnotesize
\caption{Overview of models in our study. We select a representative group of models by including models with different architectures, objectives, MIM modes, and positional encoding in our experimental setup.}
\label{tab:models}
\begin{tabular}{llllll}
\toprule
\textbf{Model} & \textbf{Arch.} & \textbf{Objective} & \textbf{MIM mode} & \textbf{Pos. enc.} \\
\midrule
DINOv2   & ViT-B/14 & MIM + CL & Latent & Add. \\
DINOv3   & ViT-B/16 & MIM + CL & Latent & RoPE \\
iBOT     & ViT-B/16 & MIM + CL & Latent & Add.  \\
MAE      & ViT-B/16 & MIM & Pixel & Add. \\
CAPI     & ViT-L/14 & MIM &  Latent & Add. \\
I-JEPA   & ViT-H/14 & MIM &  Latent & Add. \\
DINO     & ViT-B/16 & CL &  NA & Add. \\
DeiT3    & ViT-B/16 & Supervised &  NA & Add. \\
\midrule
\end{tabular}
\end{table}

\def\tmp{}%
\foreach \ds/\dsp in {%
dinov2\string_base/DINOv2,
dinov3\string_base/DINOv3,
mae\string_base/MAE,
dino\string_base/DINO
}{%
    \xappto{\tmp}{\dsp&}%
    \foreach \i in {0,1,...,4}{%
        \xappto{\tmp}{%
          \noexpand\includegraphics[width=\noexpand\linewidth]{figures/activations/\ds_out/real_0\i.png}&%
          \noexpand\includegraphics[width=\noexpand\linewidth]{figures/activations/\ds_out/synth_0\i.png}%
        }%
        \ifnum\i<4\relax%
            \gappto{\tmp}{&}%
        \else%
            \gappto{\tmp}{\\}%
        \fi%
    }%
}%

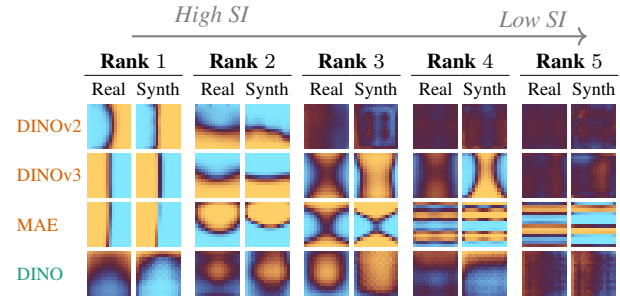
\begin{figure}[tb]
    \centering
    
    \setlength{\tabcolsep}{1pt}
    \begin{tblr}[expand=\tmp]{
        width=\linewidth,
        colspec={Q[l,m]*{10}{Q[h,c,wd=.07\linewidth]}},
        column{1}={font=\scriptsize},
        stretch=0,
        colsep=1pt,
        rowsep=1pt,
        row{1} = {abovesep=1pt, belowsep=2pt},
        row{2} = {font=\scriptsize\bfseries, belowsep=2.5pt},
        row{3} = {font=\scriptsize, abovesep=2.5pt},
        cell{3-6}{1}={fg=Dark2-B},
        cell{7-8}{1}={fg=Dark2-A},
        column{3,5,7,9,11} = {rightsep=5pt},
    }

        & \SetCell[c=11]{c}{%
            \begin{tikzpicture}[
            baseline=(current bounding box.center), 
            x=\linewidth,
            draw=gray,
            text=gray,
            ]
                \draw[->, thick] (0,0) -- (0.85,0);
                \node[above]  at (0.15,0)  {\small \textit{High SI}};
                \node[above] at (0.75,0) {\small \textit{Low SI}};
            \end{tikzpicture}
        } \\
        &
        \SetCell[c=2]{c}\footnotesize $\text{Rank}\;1$ &&
        \SetCell[c=2]{c}\footnotesize $\text{Rank}\;2$ &&
        \SetCell[c=2]{c}\footnotesize $\text{Rank}\;3$ &&
        \SetCell[c=2]{c}\footnotesize $\text{Rank}\;4$ &&
        \SetCell[c=2]{c}\footnotesize $\text{Rank}\;5$ && \\
        \cmidrule[r]{2-3}
        \cmidrule[r]{4-5}
        \cmidrule[r]{6-7}
        \cmidrule[r]{8-9}
        \cmidrule[r]{10-11}
        & Real & Synth &
        Real & Synth &
        Real & Synth &
        Real & Synth &
        Real & Synth \\
        \tmp%
    \end{tblr}
    \caption{Semantic Invariance (SI) scores reveal positional bias in top-ranked principal components.
    \textbf{Real} and \textbf{Synth} columns show activations for real and synthetic images, respectively.
    The top two components encode left/right and top/bottom biases (see \cref{fig:tblr_response}), which diminish in lower-ranked components.
    \textcolor{Dark2-B}{MIM models} exhibit clear semantic invariance, whereas the \textcolor{Dark2-A}{non-MIM} model (DINO) does not.
    See \cref{fig:activations_bigfig} for additional models and examples.}
    \label{fig:activations_small}
\end{figure}

\begin{figure*}[t]
\centering
\begin{tikzpicture}[scale=1, every node/.style={transform shape}]

\begin{groupplot}[
    group style={
        group size=2 by 1,
        horizontal sep=0.1\textwidth,
        ylabels at=edge left,
    },
    trim axis group left,
    trim axis group right,
    width=0.49\textwidth, height=5.0cm,
    ylabel near ticks,
    xlabel style={font=\small, yshift=-16pt},
    ylabel style={font=\small},
    y tick label style={font=\footnotesize},
    x tick label style={font=\footnotesize},
    ylabel={SI Score},
    cycle multi list={
        solid, dashed\nextlist
        [6 of]Dark2
    },
    enlargelimits=false,
    axis on top,
]

\nextgroupplot[
    xlabel={(a) All principal components ranked by SI.},
    xticklabel=\empty,
    ymin=0.62, ymax=1.0,
    ytick={0.65,0.75,0.85,0.95},
]

\foreach \model/\label in {
    mae_base/MAE,
    dinov3_base/DINOv3,
    ibot_base/iBOT,
    dinov2_base/DINOv2,
    ijepa_huge/IJEPA,
    capi_large/CAPI,
    deit3_base/DeiT3,
    dino_base/DINO%
} {
    \def\denom{768}%
    \def\test{capi_large}%
    \ifx\model\test
        \def\denom{1024}%
    \fi
    \def\test{ijepa_huge}%
    \ifx\model\test
        \def\denom{1280}%
    \fi
    \addplot+[thick] table [x expr={\thisrow{x}/\denom}, y=y] {figures/plots/binary_responses/\model_scores_sorted.txt};
}

\nextgroupplot[
    xlabel={(b) Top 10 principal components.},
    ylabel={},
    xmin=0, xmax=10,
    ymin=0.62, ymax=1.0,
    xticklabel=\empty,
    xtick=data,
    ytick={0.65,0.75,0.85,0.95},
    legend to name=combinedlegend,
    legend style={
        draw=none,
        font=\scriptsize,
        text=black,
        /tikz/every even column/.append style={column sep=15pt},
        legend columns=8,
    },
]

\begin{scope}
  \clip (axis cs:0,0.75) rectangle (axis cs:4.9,1.0);
  \shade[
    left color=red!15,
    right color=red!0,
    shading angle=90,
    draw=none,
  ] (axis cs:0,0.75) rectangle (axis cs:5,1.0); 
\end{scope}

\begin{scope}
  \clip (axis cs:0,0.62) rectangle (axis cs:4.9,0.75);
  \shade[
    left color=green!15,
    right color=green!0,
    shading angle=90,
    draw=none,
  ] (axis cs:0,0.62) rectangle (axis cs:5,0.75);
\end{scope}

\draw[gray, thick, dotted, opacity=0.5] ({rel axis cs:0,0} |- {axis cs:0,0.75}) -- ({rel axis cs:1,0} |- {axis cs:0,0.75});

\foreach \model/\label in {
    dinov2_base/DINOv2,
    dinov3_base/DINOv3,
    ibot_base/iBOT,
    mae_base/MAE,
    capi_large/CAPI,
    ijepa_huge/IJEPA,
    dino_base/DINO,
    deit3_base/DeiT3%
} {
    \addplot+[thick] table [x=x, y=y, restrict x to domain=0:10] {figures/plots/binary_responses/\model_scores_sorted.txt};
    \expandafter\addlegendentry\expandafter{\label}
}

\end{groupplot}

\begin{scope}
  \coordinate (a1L) at ([yshift=6pt]group c1r1.north west);
  \coordinate (a1R) at ([yshift=6pt]group c1r1.north east);

  \draw[->, thick, gray] (a1L) -- (a1R);

  \node[anchor=south west, gray] at (a1L) {\small \textit{High SI}};
  \node[anchor=south east, gray] at (a1R) {\small \textit{Low SI}};
\end{scope}

\begin{scope}
  \coordinate (a2L) at ([yshift=6pt]group c2r1.north west);
  \coordinate (a2R) at ([yshift=6pt]group c2r1.north east);

  \draw[->, thick, gray] (a2L) -- (a2R);

  \node[anchor=south west, gray] at (a2L) {\small \textit{High SI}};
  \node[anchor=south east, gray] at (a2R) {\small \textit{Low SI}};
\end{scope}

\node at ($(group c1r1.south)!0.5!(group c2r1.south)$) [below=2pt] {\ref{combinedlegend}};

\end{tikzpicture}
\caption{
Semantic invariance (\score) score in descending order. 
All scores are shown in the left plot, while the right focuses on the top 10 semantically invariant scores. 
Note that all MIM-models have a max-score $\geq 0.75$, while all non MIM models have a lower score.
}
\label{fig:invariance}
\end{figure*}
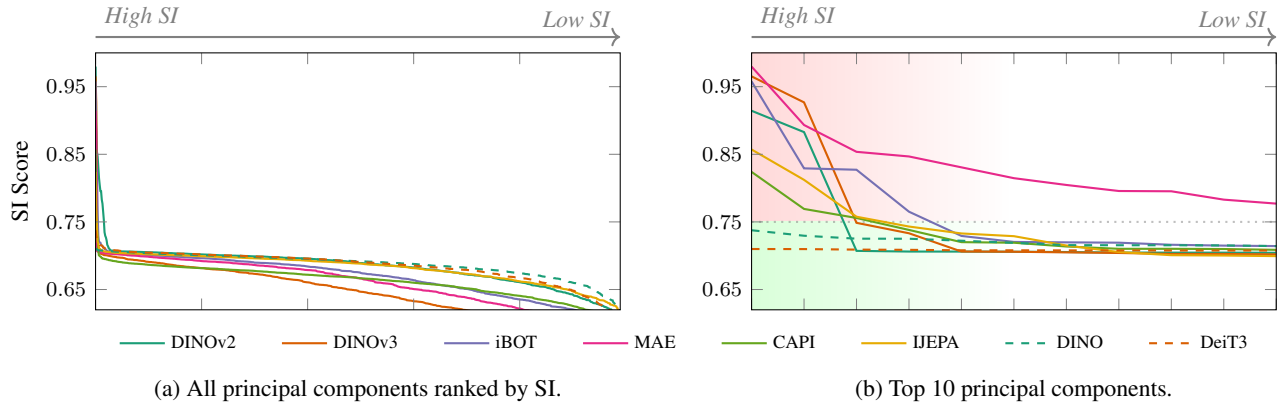

\subsection{Analyzing Information Content in SSL Tokens} \label{sec:analysis}

We examine the nature of the non-semantic information revealed by the SI-score for each model. To do so, we aggregate the activation maps for each component over patch embeddings from ImageNet validation and generated synthetic images.
\cref{fig:activations_small} visualizes the averaged activations of the principal components with the highest \score-scores, where we selected a few models for demonstration.
Extended results for all models in our study is included in \cref{fig:activations_bigfig}.
Across all MIM-based models, we observe strong positional bias in the form of left/right and top/bottom alignment (first two columns). In contrast, DINO and DeiT3 do not exhibit the same structured positional patterns.
This empirically supports our claim that MIM objectives amplify the encoding of positional information in patch tokens.
Notably, this phenomenon is present in both MIM-based models with standard additive positional embeddings (DINOv2, iBOT, CAPI, IJEPA) and those that inject positional information via RoPE in the attention mechanism (DINOv3).
We also note that the two components with the highest semantic invariance are the exact same components identified as encoding positional noise in \cref{fig:tblr_response} for all MIM models, demonstrating that the \score-score indeed captures positional noise.

\begin{figure}[tb]
\centering
\begin{tikzpicture}
\begin{axis}[
    ymin=0.62, ymax=1,
    ylabel={max SI-score},
    xlabel style={font=\footnotesize}, 
    ylabel near ticks,
    ylabel style={font=\small}, 
    y tick label style={font=\footnotesize},
    ytick={0.65,0.75,0.85,0.95},
    symbolic x coords={DINO, DeiT3, CAPI, IJEPA, DINOv2, iBOT, DINOv3, MAE},
    xtick={DINO, DeiT3, CAPI, IJEPA, DINOv2, iBOT, DINOv3, MAE},
    x tick label style={yshift=-0.15cm, rotate=45, anchor=east, font=\footnotesize},
    width=\linewidth,
    height=5.0cm,
    legend style={
      font=\footnotesize,
      at={(.99,0.05)},
      anchor=south east,
      cells={anchor=west},
      legend columns=2,
      inner sep=1pt,
    },
]

\addplot+[ycomb, mark=*, mark size=1.5pt, Dark2-A, mark options={Dark2-A}] 
    coordinates {
        (DINO, 0.7377)
        (DeiT3, 0.7099)
    };
\addlegendentry{Non-MIM}

\addplot+[ycomb, mark=triangle*, mark size=2pt, Dark2-B, mark options={Dark2-B}] 
    coordinates {
        (CAPI, 0.8239)
        (IJEPA, 0.8571)
        (DINOv2, 0.9144)
        (iBOT, 0.9578)
        (DINOv3, 0.9652)
        (MAE, 0.9797)
    };
\addlegendentry{MIM}

\draw[gray, thick, dotted, opacity=0.5] ({rel axis cs:0,0} |- {axis cs:DINO,0.75}) -- ({rel axis cs:1,0} |- {axis cs:DINO,0.75});

\foreach \x in {DINO, DeiT3, CAPI, IJEPA, DINOv2, iBOT, DINOv3, MAE} {
    \edef\tmp{\noexpand\coordinate (\x) at ({axis cs:\x,0} |- {rel axis cs:0,0});}
    \tmp
}
\end{axis}

\end{tikzpicture}
\caption{Models with MIM objectives exhibit higher max semantic invariance (SI) than models with other objectives. 
}
\label{fig:invariance_scores}
\end{figure}
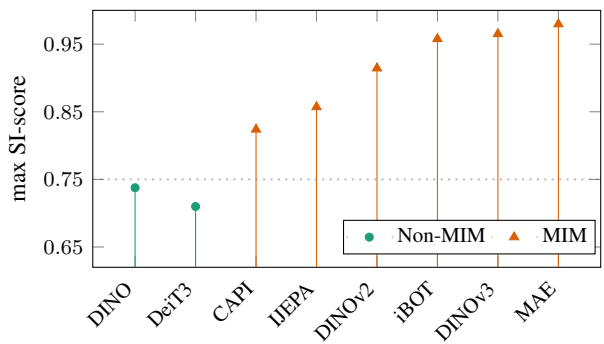

Next, we take a closer look at the \score-scores for the principal components of all the models---\cref{fig:invariance} shows the scores in descending order.
We observe that MIM-models exhibit much higher maximum \score-scores than models trained without MIM. In particular, at least two components have an \score-score higher than $0.75$ for all MIM-models. Meanwhile, models trained without MIM score below this threshold for all components, as
demonstrated by the red-green gradient.

To highlight the discrepancy between models trained with and without MIM objectives, we look at the maximum \score-score for each model in \cref{fig:invariance_scores}. This reflects the level of non-semantic noise encoded by each model, and shows a prominent gap between non-MIM and MIM-based models.

\begin{figure}[tb]
\centering%
\begin{tikzpicture}%
\begin{axis}[
    width=0.8\linewidth,
    height=5.0cm,
    xlabel={ViT layer},
    ylabel={max \score-score},
    xlabel style={font=\small},
    ylabel style={font=\small},
    y tick label style={font=\footnotesize},
    x tick label style={font=\footnotesize},
    ytick={0.65,0.75,0.85,0.95},
    xmin=0,
    xmax=11,
    ymin=0.7,
    ymax=1.01,
    cycle list={
        solid, Dark2-A\\  
        solid, Dark2-B\\  
        solid, Dark2-C\\  
        solid, Dark2-D\\  
        dashed, Dark2-A\\ 
        dashed, Dark2-B\\ 
    },
    legend style={
        cells={anchor=west},
        font=\scriptsize,
        at={(1.02,1)},
        anchor=north west,
        draw=none,
    },
    ]
    \foreach \model/\label in {
        dinov2_base/DINOv2,
        dinov3_base/DINOv3,
        ibot_base/iBOT,
        mae_base/MAE,
        dino_base/DINO,
        deit3_base/DeiT3%
    } {
        \edef\tmp{%
            \noexpand\addplot+[thick, no markers] table [x=layer, y=score, col sep=comma] {figures/plots/SI_scores_per_layer/\model_SIscores.csv};
            \noexpand\addlegendentry{\label}
        }
        \tmp
        
    }
\end{axis}%
\end{tikzpicture}%
\caption{Maximum semantic invariance score for MIM models (solid lines) and non-MIM models (dashed lines) vs. model depth. Critically, MIM models show high \score-scores in the last layers. This can be explained by the MIM objective encouraging positional information in the patch embeddings of deeper layers.}%
\label{fig:SI_per_vitlayer}%
\end{figure}
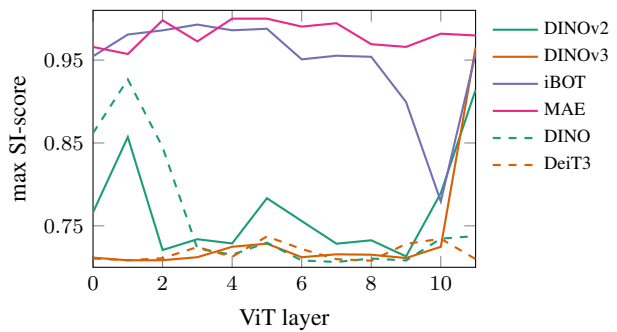

We also probe semantic invariance over model depth. \cref{fig:SI_per_vitlayer} shows the maximum \score-score per layer for a selection of the models. 
We observe that semantic invariance starts out high in the early layers, which is expected for models with additive positional embeddings, although DeiT3 is a somewhat surprising exception to this.
Specific to the MIM-based models is that the \score-score increases again for the last layers. This can be explained as the model saturates more positional information into the embeddings in preparation for solving the MIM task.
For MAE however, the score remains high across all layers.

To summarize, we confirm that the most semantically invariant components revealed by the SI score encode positional information, we show that MIM-based models consistently exhibit higher \score-scores, and we observe that the non-semantic noise either increases in the last transformer layer or is high throughout the MIM-based models.

\subsection{Cleaning with \METHODNAME for Zero-Shot Performance}
\label{sec:downstream-performance}

\begin{table*}[t]
    \footnotesize
    \centering
    \caption{Zero-shot salient segmentation with TokenCut. We evaluate on ECSSD~\citep{ecssd}, DUTS~\citep{duts}, and DUT-OMRON~\citep{dutomron}. Correcting the embeddings with \METHODNAME improves results for all MIM-based models.}
    \begin{tblr}{
      colspec={llccccccccc},
      stretch=0,
      row{3}={bg=black!10},
      row{11}={bg=highlight}
    }
    \toprule
    \SetCell[c=2]{c} && \SetCell[c=3]{c} ECSSD &&& \SetCell[c=3]{c} DUTS &&&  \SetCell[c=3]{c} DUT-OMRON \\
    \cmidrule[r]{3-5} 
    \cmidrule[r]{6-8} 
    \cmidrule[r]{9-11} 
    Pretrain & Model &
    {$\max F_\beta$} & {IoU} & {Acc.} & 
    {$\max F_\beta$} & {IoU} & {Acc.} & 
    {$\max F_\beta$} & {IoU} & {Acc.} \\
    \midrule
    \SetCell[c=11]{l} \textit{Original embeddings} &&&&&&&&&& \\
    DINO & ViT-B16 & 
        82.580 & 74.325 & 90.929 & 
        83.932 & 75.769 & 89.705 &
        59.289 & 52.851 & 83.019 \\
    DINOv2 & ViT-B16 & 
        71.319 & 63.937 & 83.147 
        & 69.701 & 63.064 & 78.891
        & 45.923 & 39.643 & 75.067 \\
    DINOv3 & ViT-B16 & 
        36.975 & 29.122 & 52.953
        & 39.874 & 31.302 & 52.264
        & 19.656 & 15.623 & 46.258 \\
    iBOT & ViT-B16 & 
        62.873 & 56.248 & 78.785 & 
        66.353 & 59.657 & 77.987 &
        33.731 & 29.602 & 67.883 \\
    CAPI & ViT-L14 & 
        72.456 & 66.083 & 84.334
        & 68.148 & 61.913 & 78.634
        & 49.150 & 42.423 & 77.762\\
    MAE & ViT-B16 & 
        79.952 & 71.067 & 89.410 & 
        79.229 & 70.227 & 86.078 &
        54.758 & 45.630 & 78.552 \\
    I-JEPA & ViT-H14 & 
        37.670 & 27.989 & 68.898 & 
        33.008 & 24.890 & 63.592 & 
        33.345 & 24.340 & 71.343 \\
    \midrule
    \SetCell[c=11]{l} \textit{Corrected embeddings} &&&&&&&&&& \\
    DINOv2 & ViT-B16 & 
        80.633 & 72.559 & 88.687
        & 84.387 & 76.785 & 89.440
        & 50.610 & 43.472 & 71.762 \\
    DINOv3 & ViT-B16 & 
        42.633 & 33.742 & 61.975 & 
        47.624 & 39.057 & 63.033 &
        23.329 & 17.485 & 51.390 \\
    iBOT & ViT-B16 & 
        66.557 & 60.167 & 78.340 & 
        72.192 & 65.618 & 80.595 &
        36.330 & 31.991 & 63.552 \\
    CAPI & ViT-L14 & 
        \textbf{85.219} & \textbf{78.084} & \textbf{92.600} & 
        \textbf{85.710} & \textbf{78.439} & \textbf{91.906} &
        59.872 & \textbf{51.315} & 80.291 \\
    MAE & ViT-B16 & 
        82.094 & 72.118 & 91.444 & 
        82.877 & 72.293 & 90.107 &
        \textbf{59.931} & 48.297 & \textbf{82.974} \\
    I-JEPA & ViT-H14 & 
        40.239 & 31.162 & 71.406 & 
        33.371 & 25.922 & 65.001 &
        35.472 & 27.038 & 76.841 \\
    \bottomrule
    \end{tblr}%
    \label{tab:salient}
\end{table*}

We use \METHODNAME to correct for semantically invariant components in local embeddings, 
and find that this improves performance in zero-shot downstream tasks for all MIM models.
Note that since non-semantic noise can be identified through linear methods, 
evaluation using learnable heads is unsuitable as they can adapt to unintentionally mask the issue.
See \cref{sec:lin-eval} for further details.

Based on the observations from \cref{sec:analysis}, we let $\mu$ be the number of components with an \score-score above a threshold of $0.75$,
and use a sharper cut-off $\tau=0.05$ to reduce suppression of the remaining components.
This makes the suppression more sparse and retains all but the most semantically invariant components.
In the case that no components have an \score-score higher than $0.75$, we set all weights $w_d=0$; this is equivalent to no suppression.

\paragraph{Salient segmentation.}

We select TokenCut~\citep{wang2023tokencut} for zero-shot evaluation of relative saliency information present in local embeddings, and follow their evaluation on three datasets---ECSSD~\citep{ecssd}, DUTS~\citep{duts}, and DUTOMRON~\citep{dutomron}---reporting intersection over union (IoU), accuracy, 
and the F-measure\footnote{
F-measure is a standard metric used in saliency detection, defined by 
$$F_\beta=\frac{(1+\beta^2)\, \text{Precision}\times\text{Recall}}{\beta^2 \, \text{Precision}+\text{Recall}}.$$ 
Prediction and recall are defined by the binary prediction mask and the ground truth. We report the max value of 255 uniformly distributed binarization thresholds, denoted max$F_\beta$. IoU, Accuracy, and max$F_\beta$ are all computed following \citet{wang2023tokencut}.
}.

\cref{tab:salient} shows that correcting the patch embeddings, in addition to using salient principal components as guides to foreground selection, can significantly boost performance.
We observe that DINO performs better out-of-the-box compared to the models with MIM objectives.
We believe this is because DINO is trained with a global objective only. 
Positional information, and to a degree local semantic information, is not as useful for the salient segmentation task which relies on global (class specific) correlations.
However, after suppressing semantically invariant components, most MIM models perform on par or better than DINO. Some notable exceptions are I-JEPA and DINOv3, which perform poorly in general for this task.

\begin{figure}[tb]
    \centering
    \setlength{\tabcolsep}{1pt}
    \footnotesize
    \begin{tabular}{cccc}
        Input & Raw pred. & \METHODNAME\ pred. & Ground Truth \\
        
        \includegraphics[width=0.22\linewidth, trim={50 0 50 0}, clip]{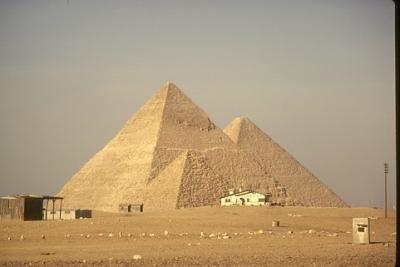} & 
        \includegraphics[width=0.22\linewidth, trim={50 0 50 0}, clip]{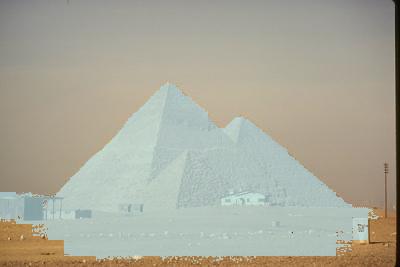} & 
        \includegraphics[width=0.22\linewidth, trim={50 0 50 0}, clip]{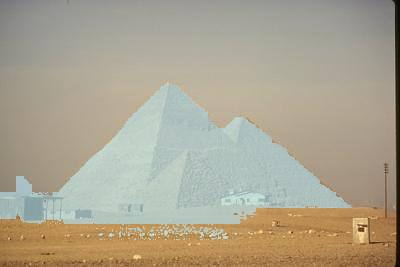} & 
        \includegraphics[width=0.22\linewidth, trim={50 0 50 0}, clip]{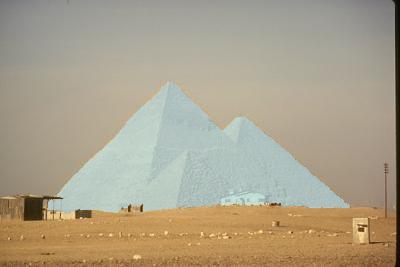} \\

        \includegraphics[width=0.22\linewidth, trim={0 50 0 100}, clip]{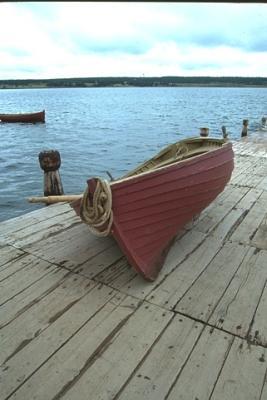} & 
        \includegraphics[width=0.22\linewidth, trim={0 50 0 100}, clip]{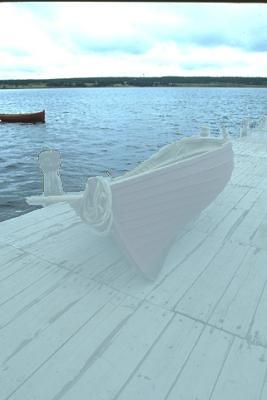} & 
        \includegraphics[width=0.22\linewidth, trim={0 50 0 100}, clip]{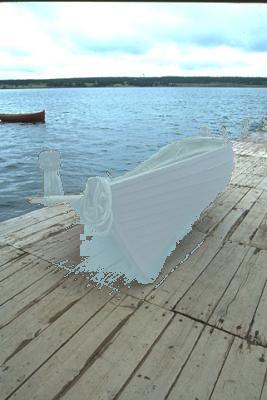} & 
        \includegraphics[width=0.22\linewidth, trim={0 50 0 100}, clip]{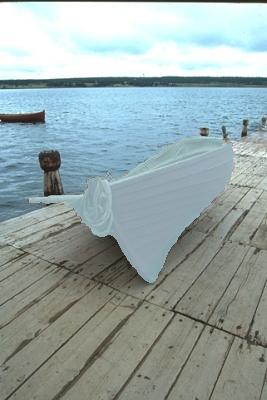} \\

        \includegraphics[width=0.22\linewidth, trim={0 120 0 30}, clip]{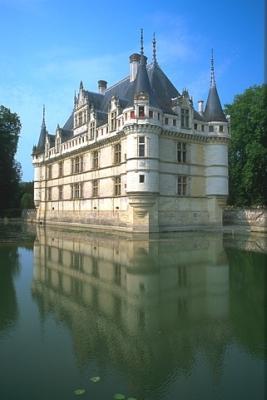} & 
        \includegraphics[width=0.22\linewidth, trim={0 120 0 30}, clip]{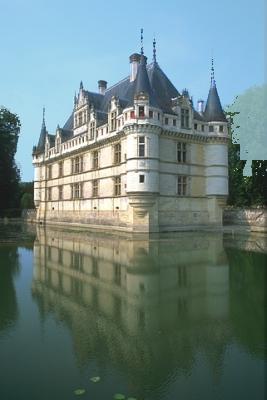} & 
        \includegraphics[width=0.22\linewidth, trim={0 120 0 30}, clip]{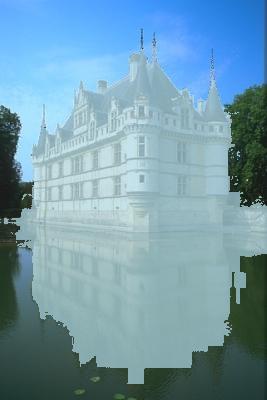} & 
        \includegraphics[width=0.22\linewidth, trim={0 120 0 30}, clip]{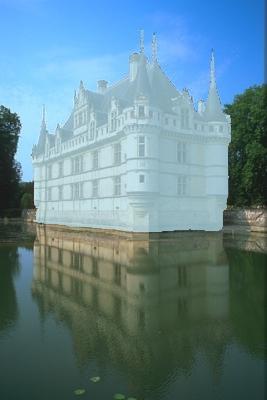} \\
    
    \end{tabular}
    \caption{Examples of salient segmentation using TokenCut~\citep{wang2023tokencut} with frozen CAPI~\citep{darcet2025cluster} on the raw embeddings (Raw pred.) and after correcting with \METHODNAME (\METHODNAME pred.). Suppressing positional noise with \METHODNAME improves the zero-shot saliency maps.
    In the last row, TokenCut is unable to generate a salient prediction from the raw embeddings, yet it succeeds when provided with \METHODNAME embeddings.
    }
    \label{fig:salient_improvement_examples}
\end{figure}

\paragraph{kNN segmentation.}
We evaluate zero-shot segmentation on ADE20k~\citep{zhou2017sceneADE20k, zhou2019semanticADE20k} by performing per-patch k-nearest neighbors (kNN) and upsampling the predictions to full image resolution using nearest neighbor interpolation.
\cref{tab:seg} shows that correcting the patch embeddings boosts performance for all models.
Additional details and results for PascalVOC~\citep{pascalvoc} are given in \cref{sec:knnseg-details}.

\begin{table}[tb]
    \footnotesize
    \centering
    \caption{
    kNN segmentation on ADE20k \cite{zhou2019semanticADE20k} reporting mean IoU and pixel accuracy (Acc).
    Correcting with \METHODNAME improves results.
    }
    \begin{tblr}{
        colspec={llcccccc},
        colsep=2pt,
        stretch=0,
        cell{1}{5-6}={bg=highlight},
        column{6,8}={font=\tiny, fg=green!65!black, leftsep=0pt},
    }
    \toprule
    & & \SetCell[c=2]{c} Original embeddings && \SetCell[c=4]{c} Corrected embeddings & \\
    \cmidrule[lr]{3-4} 
    \cmidrule[lr]{5-8} 
    Pretrain & Model &
    {IoU} & {Acc}  & 
    \SetCell[c=2]{}IoU && \SetCell[c=2]{} Acc & \\
    \midrule
    DINOv2 & ViT-B16 & 
         40.253 & 74.603 & 
         40.808 & $\uparrow$\,0.556  & 74.723 & $\uparrow$\,0.120 \\
    DINOv3 & ViT-B16 & 
         43.849 & 77.943 & 
         44.575 & $\uparrow$\,0.725  & 78.101 & $\uparrow$\,0.158 \\
    iBOT & ViT-B16 & 
         27.726 & 70.859 & 
         28.426 & $\uparrow$\,0.700  & 71.262 & $\uparrow$\,0.403 \\
    CAPI & ViT-L14 & 
         31.382 & 71.626 & 
         31.637 & $\uparrow$\,0.255  & 71.770 & $\uparrow$\,0.144  \\
    MAE & ViT-B16 & 
         11.882 & 58.002 & 
         12.651 & $\uparrow$\,0.769  & 58.592 & $\uparrow$\,0.591 \\
    I-JEPA & ViT-H14 & 
         20.952 & 60.273 & 
         21.258 & $\uparrow$\,0.306  & 60.287 & $\uparrow$\,0.014 \\
    DINO & ViT-B16 &
         21.208 & 66.478 & 
         21.208 & \SetCell{fg=gray} \,\,0.000 & 66.478 & \SetCell{fg=gray} \,\,0.000 \\
    \bottomrule
    \end{tblr}
    \label{tab:seg}
\end{table}

\paragraph{kNN classification.}

To probe the degree to which instance information is encoded in the patch embeddings, we perform classification by weighted aggregation of patch-level kNN predictions.
We use \texttt{cls}-attention weighted aggregation, except for the models trained without an instance objective (CAPI, MAE, I-JEPA), where we aggregate weighted by patch-prediction entropy instead. 
More details are given in \cref{sec:evaluation_details}.
We compare top-1 and top-5 accuracies on ImageNet~\citep{imagenet}.
The results in \cref{tab:knncls-weighted} show improvements after correcting for invariant components, although the benefit is in general less than for dense tasks.
This is expected, as classification is not as reliant on local semantics and is therefore less affected by positional noise.

\subsection{Ablations}
\label{sec:ablations}

\begin{figure}[t]
    \centering%
    \begin{tikzpicture}
    \begin{axis}[
    footnotesize,
    width=4.5cm,
    height=4.5cm,
    axis equal image,
    tick label style={font=\footnotesize},
    symbolic x coords={Caltech256,COCO,CUB,PascalVOC2012,IN1k},
    symbolic y coords={Caltech256,COCO,CUB,PascalVOC2012,IN1k},
    xtick=data,
    ytick=data,
    xticklabel style={rotate=45, anchor=east},
    xmin=Caltech256, xmax=IN1k,
    ymin=Caltech256, ymax=IN1k,
    point meta max=0.01,
    point meta min=0,
    enlargelimits=true,
    no markers,
    colormap/viridis,
    colorbar right,
    every colorbar/.append style={
      title={$d_{\cos}$ },
      title style={font=\footnotesize, yshift=0.1cm},
      width=0.25cm,
      xshift=0.5cm,
      yticklabel style={
        align=right,
        font=\footnotesize,
        /pgf/number format/.cd,
        fixed,
        precision=1,
        fixed zerofill,
        /tikz/.cd,
      },
    },
  ]
    \addplot+[matrix plot*, point meta=explicit] table [x=x, y=y, meta=z, col sep=comma] {figures/plots/SI_scores_cosdistance_vs_dataset.csv};
        
    \end{axis}%
    \end{tikzpicture}%
    \caption{Cosine distance between \score-scores for DINOv2 using various datasets for semantically informative images. 
    The low values in $(.0025,.0032)$ indicate that the \score-score is consistent across datasets.}
    \label{fig:SI_vs_dataset}%
\end{figure}
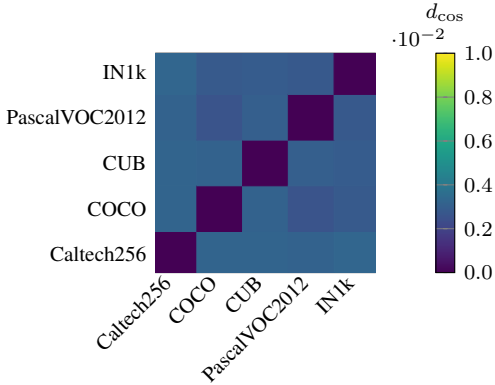

\paragraph{SI-score sensitivity to dataset choice.}
To probe the \score-score sensitivity to dataset choice, we ablate over Caltech256~\citep{caltech256}, COCO-Stuff164k~\citep{coco}, CUB200~\citep{cub200}, PASCAL VOC~\citep{pascalvoc}, and ImageNet~\citep{imagenet} using DINOv2 as the backbone. 
We calculate the cosine distance between the score vectors over the $D=768$ components. 
We observe that the distances are low (between $0.0025$ and  $0.0032$), which means that the scores for each component remain consistent despite using different datasets with different distributions and of various sizes to instantiate semantically informative input.
This indicates that the \score-score is not sensitive to dataset choice.
We show the \score-score distance for each pair of datasets in \cref{fig:SI_vs_dataset}.

\paragraph{Use of scaling function.}
We ablate the necessity of filtering the SI-scores with the scaling function from \cref{eq:fermi},
by evaluating on kNN classification of the average pooled patch embeddings on ImageNet~\citep{imagenet}.
The details of the evaluation method are given in \cref{sec:cls-avg-details}.
The results in \cref{tab:fermi-knn-ablation} show that applying \METHODNAME without scaling the SI-scores can reduce performance.
Evaluating on salient segmentation shows the same result; see \cref{tab:fermi-salient-ablation}.
We posit that this is because too many components are suppressed when the scores are not scaled in the projection, resulting in detrimental information loss.
While the SI-score allows us to identify the most semantically invariant components, the raw scores remain too high for the remaining components, unnecessarily dampening their contribution.

\begin{table}[tb]
    \footnotesize
    \centering
    \caption{Weighted kNN classification on ImageNet \cite{imagenet} by aggregating patch predictions. We compare original vs. \METHODNAME-corrected embeddings across backbones.
    }
    \begin{tblr}{
        colspec={llcccccc},
        colsep=2pt,
        stretch=0,
        cell{1}{5-6}={bg=highlight},
        column{6,8}={font=\tiny, fg=green!65!black, leftsep=0pt},
    }
    \toprule
    & & \SetCell[c=2]{c} Original embeddings && \SetCell[c=4]{c} Corrected embeddings & \\
    \cmidrule[lr]{3-4} 
    \cmidrule[lr]{5-8} 
    Pretrain & Model &
    {Acc@1} & {Acc@5}  & 
    \SetCell[c=2]{}Acc@1 && \SetCell[c=2]{} Acc@5 & \\
    \midrule
    DINOv2 & ViT-B16 & 
         82.32 & 96.29 & 
         82.59 & $\uparrow$\,0.27  & 96.30 & $\uparrow$\,0.01  \\
    DINOv3 & ViT-B16 & 
         81.47 & 95.57 & 
         81.48 & $\uparrow$\,0.01  & 95.59 & $\uparrow$\,0.01   \\
    iBOT & ViT-B16 & 
         71.45 & 90.03 & 
         71.60 & $\uparrow$\,0.15  & 90.10 & $\uparrow$\,0.07 \\
    CAPI\smash{\textsuperscript{\dag}} & ViT-L14 & 
         70.81 & 91.09 & 
         71.25 & $\uparrow$\,0.43  & 91.33 & $\uparrow$\,0.24 \\
    MAE\smash{\textsuperscript{\dag}} & ViT-B16 & 
         59.30 & 81.49 & 
         60.62 & $\uparrow$\,1.32  & 82.38 & $\uparrow$\,0.89  \\
    I-JEPA\smash{\textsuperscript{\dag}} & ViT-H14 & 
         75.68 & 91.70 & 
         75.89 & $\uparrow$\,0.21  & 91.76 & $\uparrow$\,0.06 \\
    DINO & ViT-B16 &
         66.08 & 86.13 & 
         66.08 & \SetCell{fg=gray} \,\,0.00  & 86.13 & \SetCell{fg=gray} \,\,0.00 \\
    \bottomrule
    \end{tblr}%
    \\[1pt]
    {\scriptsize \textsuperscript{\dag}Aggregation weighted by entropy for models with no class token objective; otherwise weighted by class attention.}
    \label{tab:knncls-weighted}
\end{table}

\begin{table}[tb]
    \footnotesize
    \centering
    \caption{
    Ablation on the scaling function in \cref{eq:fermi}, evaluated on kNN classification of average pooled patch embeddings on ImageNet~\citep{imagenet}.
    }
    \resizebox{\linewidth}{!}{%
    \begin{tblr}{
        colspec={llcccc},
        colsep=2pt,
        stretch=0,
        cell{1}{5-6}={bg=highlight},
    }
    \toprule
    & & \SetCell[c=2]{c} \METHODNAME without scaling && \SetCell[c=2]{c} \METHODNAME with scaling & \\
    \cmidrule[r]{3-4} 
    \cmidrule[]{5-6} 
    Pretrain & Model &
    {kNN Acc@1} & {kNN Acc@5}  & 
    {kNN Acc@1} & {kNN Acc@5}  \\
    \midrule
    DINOv2 & ViT-B16 & 
        77.068 & 91.588 & 
        77.102 & 91.635 \\
    DINOv3 & ViT-B16 & 
        76.354 & 91.306 & 
        76.588 & 91.612 \\
    iBOT & ViT-B16 & 
        59.194 & 79.610 & 
        59.498 & 79.918 \\
    CAPI & ViT-L14 & 
        55.720 & 76.950 & 
        56.444 & 77.742 \\
    MAE & ViT-B16 & 
        47.596 & 69.108 & 
        47.758 & 69.442 \\
    I-JEPA & ViT-H14 & 
        71.422 & 86.112 & 
        71.390 & 86.168 \\
    \bottomrule
    \end{tblr}}
    \label{tab:fermi-knn-ablation}
\end{table}

\paragraph{Comparison with RASA.}
We compare \METHODNAME with RASA by correcting the patch embeddings from Franca using pretrained RASA weights from~\citepos{vekataramanan2025franca} work.
The results in \cref{tab:rasa-ablation} show that \METHODNAME yields higher performance.
We argue that the significant improvement stems from \METHODNAME correcting non-semantic noise directly, while RASA has to learn the positional cues to remove.
We also note that RASA requires training $9$ linear layers of dimension $D\times2$ ($D=768$ for ViT-B), while \METHODNAME is \textbf{simpler}, requires \textbf{no training}, and can be \textbf{attached to any model} as a single $D\times D$ linear head.

\begin{table}[t]
    \footnotesize
    \centering
    \caption{Comparing \METHODNAME with RASA on Franca ViT-B/14 for zero-shot salient segmentation on ECSSD~\citep{ecssd} and kNN classification on ImageNet~\citep{imagenet}.}
    \begin{tblr}{
        colspec={lccccc},
        stretch=0,
        colsep=2pt,
        row{5}={font=\bfseries},
    }
    \toprule
    & \SetCell[c=3]{c} ECSSD (Sal. Seg) &&& \SetCell[c=2]{c} IN1k (kNN cls.) && \\
    \cmidrule[r]{2-4} \cmidrule[l]{5-6} 
    Method & max $F_\beta$ & IoU & Acc. & Acc@1 & Acc@5 \\
    \midrule
    Franca & 71.615 & 64.899 & 83.982 & 64.920 & 85.872  \\
    Franca + RASA  & 68.220 & 68.220 & 85.935 & 64.890 & 85.886 \\
    Franca + \METHODNAME & 84.176 & 76.985 & 91.514 & 65.084 & 86.006 \\
    \bottomrule
    \end{tblr}
    \label{tab:rasa-ablation}
\end{table}

\section{Discussion}
\label{sec:discussion}
Our analysis indicates a tendency for MIM to ``cheat'' to perform the matching of masked tokens by dedicating capacity to patch location cues. 
Furthermore, \cref{fig:activations_small} shows that this happens in both latent and reconstructive MIM\@.
The results are not totally unexpected; the reconstruction objective requires positional information to perform the task to some extent.
However, we note that latent MIM models exhibit this property with or without additive positional encoding, such as for DINOv3.
The models learn to dedicate capacity to non-semantic information.
This makes sense if we consider how much this actually helps the MIM objective; by being able to correctly select the patch position in the image, the search is reduced by a ratio of $H_z \times W_z$. 
For a ViT-B/14 model, this results in a reduction of $\times256$, significantly improving the loss of the model.
While MIM improves performance on dense objectives, it does so at a \textit{cost}.
The importance and severity of this positional noise is corroborated by several works~\citep{wang2024sinder,darcet2025cluster,yang2024denoising}.
Our method shows that \textit{the problem is pervasive for MIM models, and does not meaningfully occur in non-MIM models}.

\paragraph{Limitations and Further Work.}
We restrict evaluation to raw patch embeddings; both kNN and TokenCut operate directly on the representations without additional projections or heads. 
This is a conscious choice, as further transformations would confound the effect of \METHODNAME with that of the evaluation model itself, which is beyond the scope of the current work.
An avenue for future work is to study how \METHODNAME interacts with more elaborate evaluation protocols.
Lastly, the linear assumption as a mixture of sources (\cref{sec:lindecomp}) is a simplification, but given our empirical results, this assumption seems to at least partially hold.

\section{Conclusion}
\label{sec:conclusion}
We demonstrate that masked image modeling (MIM) objectives bias vision transformers toward encoding positional noise, corroborating our hypothesis that such noise persists even in inputs devoid of semantic content.
This suggests that MIM learning signals are solved by short-cutting the intended objective of learning better local representations, and while this helps solve the pretext task, it reduces zero-shot generalization and semantic fidelity. 
To diagnose and address this issue, we introduce a Semantic Invariance Score and the lightweight, post-hoc denoising method \METHODNAME, which consistently suppresses positional noise and improves downstream performance. 
Our findings highlight a fundamental trade-off in MIM and offer practical tools for building more semantically robust self-supervised models.

\section*{Acknowledgments}
This work was funded by RCN (the Research Council of Norway) through Visual Intelligence, Centre for Research-based Innovation (309439), and in part by the RCN–NRF (National Research Foundation of Korea) joint project AURoRA (359216, RS-2025-03522980).
We acknowledge Sigma2 (Project~NN8104K) for access to the LUMI supercomputer, owned by the EuroHPC Joint Undertaking, hosted by CSC (Finland) and the LUMI consortium through Sigma2, Norway.

%% file: sec/X_suppl.tex
\clearpage
\appendix
\setcounter{page}{1}
\setcounter{section}{0}

\renewcommand\thetable{\Alph{section}.\arabic{table}}
\renewcommand\thefigure{\Alph{section}.\arabic{figure}}
\counterwithin{figure}{section}
\counterwithin{table}{section}
\counterwithin{equation}{section}

\maketitlesupplementary

\section{Self-Supervised Objectives}
\paragraph{Contrastive Learning.}
In this work, we focus on the contrastive self-distillation objective as presented in \citet{caron2021emerging}.
Given an image $x$, two random augmentations are applied yielding the views $u,v$.
The views are sent through a teacher-student framework giving $p = f_\theta (v) \in \mathbb{R}^d$ and $q = f_{\hat\theta}(u)\in \mathbb{R}^d$, where $\theta$ and $\hat{\theta}$ denote the teacher and student weights, respectively.
The loss minimizes the cross-entropy
\begin{align}
    \mathcal{L}_{\textsc{CLS}} = - q^\top \log p.    
\end{align}
Importantly, this loss is applied on the global representations, given by the \texttt{CLS} token in ViTs. The teacher and student share the same architecture, comprising a backbone and a projection head for the global representation. The teacher is updated as an exponential moving average of the student.

\paragraph{Masked Image Modeling.}
The core idea of the MIM objective is to reconstruct masked parts of an image  when given visible parts as context. While the masking strategy varies between the MIM-based methods, the general setup can be summarized as follows.
The input $x$ is obscured by a random mask $m$ and passed through an encoder to give a context $z=f \left( m(x) \right)$ with which to make a prediction $\hat s = g(z)$ about the unmasked image $x$.
The prediction can be made in the pixel space or in the latent space where the target $s$ is given by passing $x$ through the encoder.
The loss is
\begin{align}
    \mathcal{L}_{\textsc{mim}} = \ell(\hat s, s)
\end{align}
where $\ell$ is the mean squared error~\citep{he2022mae}, euclidean distance~\citep{assran2023ijepa} or negative cross entropy~\citep{oquab2024dinov2, simeoni2025dinov3, zhou2022ibot, darcet2025cluster}.
Notably, several frameworks~\citep{oquab2024dinov2, simeoni2025dinov3, zhou2022ibot} employ a combination of contrastive loss over the global \texttt{CLS}-tokens in a multi-view setup with a local MIM-based loss over masked patch tokens. 

\section{Details on Activation and Distributions}
\label{sec:appendix_detail_activation}

\begin{figure}[tb]
    \centering
    \includegraphics[width=0.3\linewidth]{figures/pos_split_examples/dinov2_base_img1_PC1_wce_overlay.png}
    \includegraphics[width=0.3\linewidth]{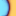}
    \caption{
    The relation of soft responses of DINOv2 to $\mathrm{PC}_1$ in \Cref{fig:tblr_response} (left), compared with the multinomial distribution of activations $P_1$ from \Cref{fig:activations_small} and \Cref{sec:seminv} (right).
    The responses are binarized, and the probability distribution $P_1$ is computed as a multinomial distribution over all tokens in the image.
    }
    \label{fig:comparison}
\end{figure}

\def\tmp{}%
\foreach \ds/\dsp in {%
dinov2\string_base/DINOv2,
dinov3\string_base/DINOv3,
ibot\string_base/iBOT,
mae\string_base/MAE,
capi\string_large/CAPI,
ijepa\string_huge/IJEPA,
dino\string_base/DINO,
deit3\string_base/DeiT3%
}{%
    \xappto{\tmp}{\dsp&}%
    \foreach \i in {0,1,...,9}{%
        \xappto{\tmp}{%
          \noexpand\includegraphics[width=\noexpand\linewidth]{figures/activations/\ds_out/real_0\i.png}&%
          \noexpand\includegraphics[width=\noexpand\linewidth]{figures/activations/\ds_out/synth_0\i.png}%
        }%
        \ifnum\i<9\relax%
            \gappto{\tmp}{&}%
        \else%
            \gappto{\tmp}{\\}%
        \fi%
    }%
}%

\begin{figure*}[tb]
    \centering
    \setlength{\tabcolsep}{1pt}
    \begin{tblr}[expand=\tmp]{
        width=1.02\linewidth,
        colspec={Q[l,m]*{20}{Q[h,c,wd=.041\linewidth]}},
        stretch=0,
        colsep=0.9pt,
        rowsep=0.9pt,
        row{1} = {abovesep=1pt, belowsep=6pt},
        row{2} = {font=\scriptsize\bfseries, belowsep=2.5pt},
        row{3} = {font=\scriptsize, abovesep=2.5pt},
        column{1}={font=\scriptsize},
        row{10} = {abovesep=4pt},
        row{9} = {belowsep=4pt},
        cell{2-9}{1}={fg=Dark2-B},
        cell{10-11}{1}={fg=Dark2-A},
        column{3,5,7,9,11,13,15,17,19} = {rightsep=3.8pt},
    }

        & \SetCell[c=21]{c}{%
            \begin{tikzpicture}[
            baseline=(current bounding box.center), 
            x=\linewidth,
            draw=gray,
            text=gray,
            ]
                \draw[->, thick] (0,0) -- (0.95,0);
                \node[above]  at (0.08,0)  {\small \textit{(Higher SI)}};
                \node[above] at (0.85,0) {\small \textit{(Lower SI)}};
                \node[above] at (0.45,0) {\small Principal components};
            \end{tikzpicture}
        } \\
    
        &
        \SetCell[c=2]{c}\footnotesize$\text{Rank}\;1$ &&
        \SetCell[c=2]{c}\footnotesize$\text{Rank}\;2$ &&
        \SetCell[c=2]{c}\footnotesize$\text{Rank}\;3$ &&
        \SetCell[c=2]{c}\footnotesize$\text{Rank}\;4$ &&
        \SetCell[c=2]{c}\footnotesize$\text{Rank}\;5$ &&
        \SetCell[c=2]{c}\footnotesize$\text{Rank}\;6$ &&
        \SetCell[c=2]{c}\footnotesize$\text{Rank}\;7$ &&
        \SetCell[c=2]{c}\footnotesize$\text{Rank}\;8$ &&
        \SetCell[c=2]{c}\footnotesize$\text{Rank}\;9$ &&
        \SetCell[c=2]{c}\footnotesize$\text{Rank}\;10$ && \\
        \cmidrule[r]{2-3}
        \cmidrule[r]{4-5}
        \cmidrule[r]{6-7}
        \cmidrule[r]{8-9}
        \cmidrule[r]{10-11}
        \cmidrule[r]{12-13}
        \cmidrule[r]{14-15}
        \cmidrule[r]{16-17}
        \cmidrule[r]{18-19}
        \cmidrule[r]{20-21}
        & Real & Synth &
        Real & Synth &
        Real & Synth &
        Real & Synth &
        Real & Synth &
        Real & Synth &
        Real & Synth &
        Real & Synth &
        Real & Synth &
        Real & Synth \\
        \tmp%
    \end{tblr}
    \caption{Distributions for the top $10$ principal components, ranked by semantic invariance (SI). 
    \textbf{Real} columns correspond to $P_d$ and \textbf{Synth} correspond with $Q_d$ from \Cref{sec:seminv}, and each column compares the activations for real images from ImageNet, and generated synthetic images.
    \textcolor{Dark2-B}{MIM models} are shown in the top rows; \textcolor{Dark2-A}{non-MIM models} in the bottom rows. \textcolor{Dark2-B}{MIM models} exhibit higher semantic invariance than the \textcolor{Dark2-A}{non-MIM models}, as seen by the clear positional bias in the activations.
    }
    \label{fig:activations_bigfig}
\end{figure*}
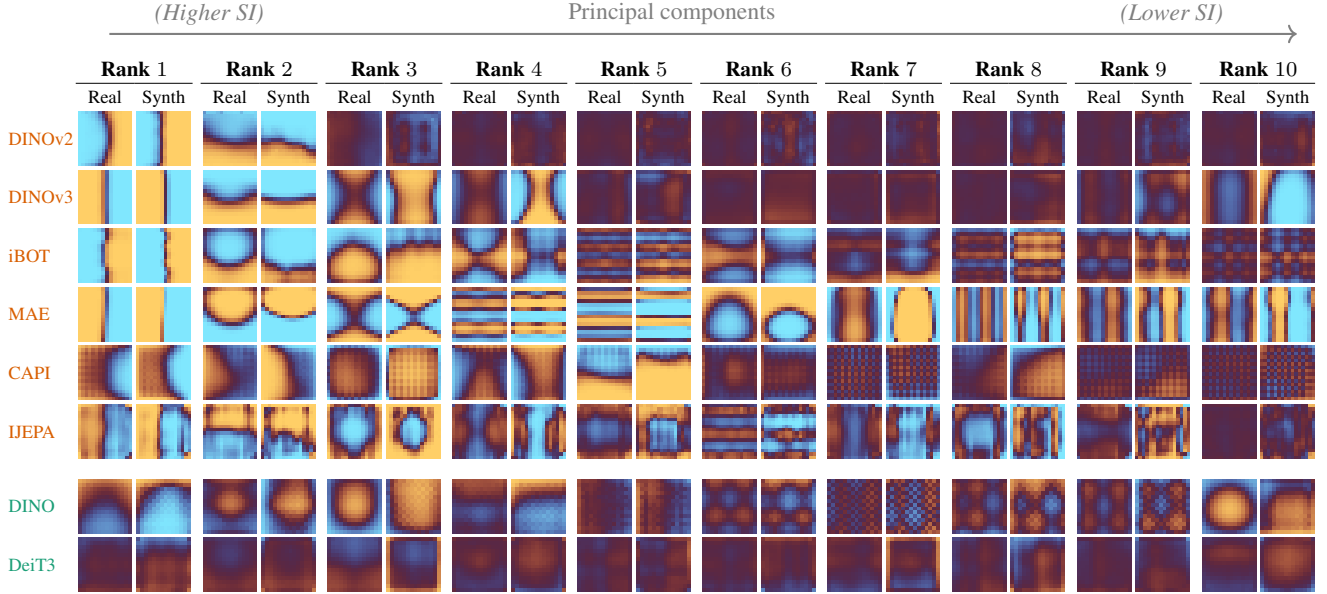

In this section, we elaborate on the details of responses, activations, and distributions. 
As explained in the main text, we compute responses for all images in the dataset for each principal component. 
We binarize the activations, and compute the token-wise empirical distributions as individual Bernoulli distributions for each token in the image, $P_{d,n}$ from \Cref{sec:seminv}.
As an ensemble, this can be taken as $P_d \sim \mathrm{Multinomial}(2,N)$, forming a full distribution over the image. 

Divergence measures would be a natural choice to compare $P_d, Q_d$, however, we find that pure divergence measures such as Jensen-Shannon yield suboptimal scores in our testing. 
If $P_{d,n}=Q_{d,n}=0.5$, a proper divergence such as the Jensen-Shannon requires that the distributions are equal, e.g., $D_{\text{JS}}(P_{d,n},Q_{d,n})=1$.
However, in this case we are unsure if individual activations actually agree or not. In other words, a proper divergence yields high scores when activation maps are highly uncertain.

In contrast, our proposed \score-score given in \cref{eq:SIscore} yields $\text{SI}(P_{d,n},Q_{d,n})=\sqrt{0.5}$ for the same example, reflecting the inherent uncertainty in agreement in the two responses, and correctly identifies similar responses between real data and the non-semantic synthetic data.

\Cref{fig:comparison} illustrates the relation between responses and empirical estimates of the distribution.
To expand on \cref{fig:activations_small} in the main article, \cref{fig:activations_bigfig} shows the top 10 principal components for each model in our study, sorted by descending SI-scores.

\subsection{Form of the \score-score}
\label{sec:vector_si}
When we calculate the \score-scores in \Cref{sec:seminv}, we do so over the multinomial distributions $P_d, Q_d$.
To clarify, we consider the inner products and norms as being taken over the \textit{support set}, not the multivariate dimensions.

Let \(P_d, Q_d \in \mathbb{R}^N\) and let \(1 \in\mathbb{R}^N\).
Define the augmented vectors
\[
\tilde P_d \;=\; \begin{bmatrix} P_d , {1}-P_d \end{bmatrix},\qquad
\tilde Q_d \;=\; \begin{bmatrix} Q_d , {1}-Q_d \end{bmatrix}
\in \mathbb{R}^{2N}.
\]
Then we calculate the scores by averaging over the multivariate dimensions, yielding
\begin{subequations}\label{eq:geoSIscore}
\begin{align}
    s_d &= \score(P_d,Q_d) \\
    &= \frac{1}{N}1^\top \Bigg(2\,\frac{\langle \tilde P_d, \tilde Q_d \rangle}{\|\tilde P_d\|_2 + \|\tilde Q_d\|_2}\Bigg),\\
    &= \frac{2}{N}\sum_n \frac{P_{d,n}Q_{d,n} + (1-P_{d,n})(1-Q_{d,n})}{\sqrt{P_{d,n}^2 + (1-P_{d,n})^2}+\sqrt{Q_{d,n}^2+(1-Q_{d,n})^2}},
\end{align}
\end{subequations}
where \(\langle\cdot,\cdot\rangle\) and \(\|\cdot\|_2\) denotes the Euclidean inner product and norm (over the support set), respectively.

\section{Generating synthetic images}
\label{sec:synth_data}

\begin{figure*}[tb]
    \centering
    \includegraphics[width=0.11\linewidth]{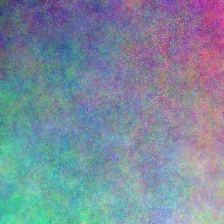}
    \includegraphics[width=0.11\linewidth]{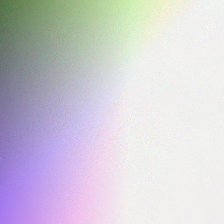}
    \includegraphics[width=0.11\linewidth]{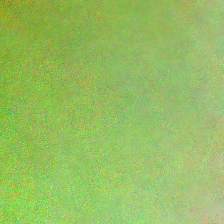}
    \includegraphics[width=0.11\linewidth]{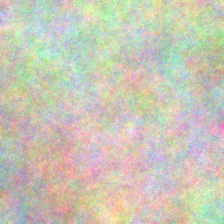}
    \includegraphics[width=0.11\linewidth]{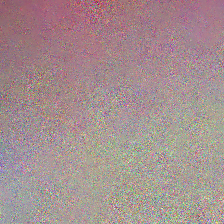}
    \includegraphics[width=0.11\linewidth]{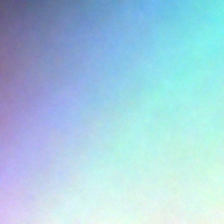}
    \includegraphics[width=0.11\linewidth]{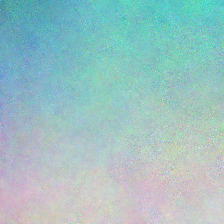}
    \includegraphics[width=0.11\linewidth]{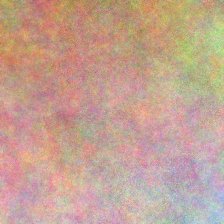}
    \caption{Examples of generated synthetic non-semantic images.}
    \label{fig:synth-data}
\end{figure*}

Let $\Omega=\{1,\dots,H\}\times\{1,\dots,W\}$ and $X\in\mathbb{R}^{C\times H\times W}$. For each image, draw mixture weights
\begin{align}
\mathbf{w}=(w_1,w_2,w_3)\sim\mathrm{Dir}(\alpha_1,\alpha_2,\alpha_3).    
\end{align}
We generate three components independently:

\begin{enumerate}[leftmargin=*, itemsep=1em]
    \item \textbf{Pink Noise}: $X_{\mathrm{pink}}$ is zero-mean with isotropic power spectrum
    $$
    \mathbb{E}\!\left[\,|\mathcal{F}\{X_{\mathrm{pink}}\}(\xi)|^2\,\right]\;\propto\;\|\xi\|^{-\beta},\qquad \xi\in\mathbb{Z}^2\setminus\{0\},
    $$
    with $\beta\approx 2$, following the power-law slope typical of natural images~\citep{simoncelli2001}.
    
    \item \textbf{Modulated White Noise}: $X_{\mathrm{white}}=M\odot W$, where $W\overset{\text{i.i.d.}}{\sim}\mathcal{N}(0,1)$ and the nonnegative modulation $M=g(P)$ is a smooth field obtained from a pink process $P$ (as above) and a bounded mapping $g$ that sets the local standard deviation. This yields a heteroscedastic Gaussian field with variance $\sigma^2(x)=M(x)^2$ and long-range variance correlations.

    \item \textbf{Gradient Field}: $X_{\mathrm{grad}}$ is a random low-degree polynomial---or equivalently, a very low-pass random field---concentrating energy near $\xi=0$.
\end{enumerate}
\vspace{5pt}
Synthesized images are then given by the convex mixture
\begin{align}
X \;=\; w_1\,X_{\mathrm{white}} \;+\; w_2\,X_{\mathrm{pink}} \;+\; w_3\,X_{\mathrm{grad}}.
\end{align}
Assuming zero mean and independence between components, the expected power spectrum of $X$ is
\begin{align}
S_X(\xi)\;=\;\mathbb{E}\!\left[\,|\mathcal{F}\{X\}(\xi)|^2\,\right]\;=\;\sum_{i=1}^3 \mathbb{E}[w_i^2]\,S_{X_i}(\xi),    
\end{align}
i.e., a convex combination of the components’ spectra.

\citet{simoncelli2001} show that natural images exhibit approximate scale invariance with a $1/\|\xi\|^{\beta}$ law in the power spectrum (amplitude $\sim 1/\|\xi\|^{\beta/2}$), together with large-scale illumination/contrast fluctuations.
In the construction above, $X_{\mathrm{pink}}$ directly imposes the $1/\|\xi\|^{\beta}$ decay, giving second-order statistics aligned with natural image ensembles. 
Meanwhile, $X_{\mathrm{grad}}$ injects additional low-frequency near-DC energy, modeling global trends and illumination. 
Finally, $X_{\mathrm{white}}$ introduces spatially varying contrast via a pink variance field, capturing long-range correlations of local variance found in natural scenes.

Thus $S_X(\xi)$ inherits a natural-image-like spectrum: a power-law falloff dominated by $X_{\mathrm{pink}}$, boosted near $\xi=0$ by $X_{\mathrm{grad}}$, and with heteroscedasticity from $X_{\mathrm{mw}}$.
We illustrate examples of synthetic images in \cref{fig:synth-data}.

During testing, we also experimented with standard probability distributions such as Gaussian or Uniform noise, in addition to simple mono-colored and gradient images, which provides similar responses to our proposed synthetic data. 
However, we considered these less appropriate given the mismatch in frequency response, which differs significantly from natural images.
Hence, we designed our synthesis to better match key properties of natural images without additional semantic content.

\section{Effect of the Scaling Function}
\begin{figure}[tb]
    \centering
    \begin{tikzpicture}
    \begin{axis}[
      width=\linewidth,
      height=0.6\linewidth,
      domain=0:2,
      declare function={
        sig(\x) = 1/(1+exp(-\x));
        f(\x,\m,\t) = sig((\m - \x)/\t)/sig(\m/\t);
        g(\x) = exp(1-\x)/exp(1);
      },
      no markers,
      samples=50,
      xtick={0,1,2},
      xticklabels={1,384,768},
      xlabel={Principal component},
      ylabel={Scaling function $t$},
      label style={font=\scriptsize},
      tick label style={font=\scriptsize},
      legend style={
        at={(0.98,0.98)}, anchor=north east, font=\scriptsize,
        draw=none, fill=none
      },
      legend cell align={left},
      ]
      \addplot+ {g(x)};
      \addlegendentry{SI scores}
      
      \addplot+ {g(x)*f(x,0.5,0.05)};
      \addlegendentry{$\mu{=}192,\ \tau{=}0.05$}
      
      \addplot+ {g(x)*f(x,1,0.1)};
      \addlegendentry{$\mu{=}384,\ \tau{=}0.1$}
      
      \addplot+ {g(x)*f(x,1.5,0.2)};
      \addlegendentry{$\mu{=}576,\ \tau{=}0.2$}
    \end{axis}
  \end{tikzpicture}
    \caption{Plot showing the effect of the scaling function, filtering with the Fermi window in \cref{eq:fermi}, on semantic invariance (SI) scores with various choices of $\mu, \tau$. The $\mu$ parameter controls how many components are filtered, while $\tau$ controls the smoothness.}
    \label{fig:weights}
\end{figure}
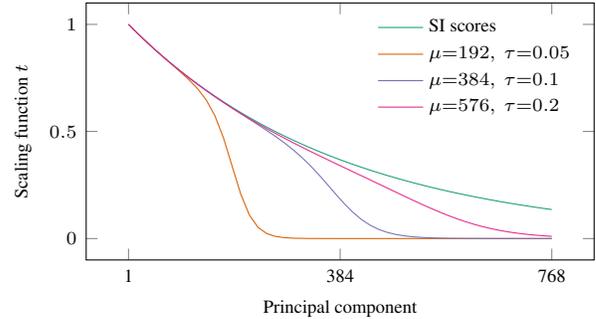

The filtering of SI scores acts as a smooth low-pass filter over the PCA spectrum. 
We selected the Fermi window as a smooth approximation of hard truncation, due to its precedence as a regularizer in image reconstruction. \Cref{fig:weights} illustrates the effect of the scaling function on the SI scores.

We ablate the effect of removing the scaling function in \METHODNAME in \Cref{sec:ablations}, showing the downstream performance of kNN classification on the aggregated patch embeddings in \Cref{tab:fermi-knn-ablation}. We provide additional results in \Cref{tab:fermi-salient-ablation} by showing the effect for salient segmentation.
We see a boost in performance by $1$--$9\%$ points for all models, except DINOv2 and CAPI, which have reduced performance by $<1\%$.
Since the majority of the \score-scores are $>0.6$ for all models, projecting the representations with \METHODNAME without scaling the \score-scores results in suppressing a majority of the principal components, reducing performance when useful information is encoded in any but the most semantic components.
We chose to include the ablation with kNN in the main paper, as the salient segmentation task requires less fine-grained information about the patch contents, making kNN more informative as an ablation task.

\begin{table}[tb]
    \footnotesize
    \centering
    \caption{
    Ablation on the scaling function in \cref{eq:fermi}, evaluated on salient segmentation for ECSSD~\citep{ecssd}. Removing the scaling function from \METHODNAME generally leads to reduced performance.
    }
    \begin{tblr}{
        colspec={lcccccc},
        colsep=2pt,
        stretch=0,
        cell{1}{5-6}={bg=highlight},
    }
    \toprule
    & \SetCell[c=3]{c} \METHODNAME without scaling &&& \SetCell[c=3]{c} \METHODNAME with scaling &&& \\
    \cmidrule[r]{2-4} 
    \cmidrule[]{5-7} 
    Pretrain &
    {max $F_\beta$} & {IoU}  & 
    {Acc.} & {max $F_\beta$} & {IoU}  & 
    {Acc.} \\
    \midrule
    DINOv2 & 
        81.615 & 73.533 & 89.472 &
        80.633 & 72.559 & 88.687 \\
    DINOv3 & 
        39.867 & 32.426 & 59.745 &
        42.633 & 33.742 & 61.975 \\
    iBOT & 
        64.702 & 58.401 & 76.419 &
        66.557 & 60.167 & 78.340 \\
    CAPI & 
        85.555 & 78.446 & 93.460 &
        85.219 & 78.084 & 92.600 \\
    MAE & 
        73.264 & 62.877 & 85.655 &
        82.094 & 72.118 & 91.444 \\
    I-JEPA & 
        32.646 & 24.327 & 68.927 &
        40.239 & 31.162 & 71.406 \\
    \bottomrule
    \end{tblr}
    \label{tab:fermi-salient-ablation}
\end{table}

\section{Evaluation Details}
\label{sec:evaluation_details}

Throughout the paper, we provide several evaluations and experiments.
In this section, we exposition some of the details for each evaluation method.

\subsection{TokenCut}\label{sec:tokencut_details}
We use the official TokenCut~\citep{wang2023tokencut} implementation with their graph cut segmentation algorithm for the patch embeddings and the bilateral solver for edge-aware post-processing to refine the segmentations up to original image size.
TokenCut is a natural extension of spectral clustering and graph cuts to token representations, where patches are either classified as belonging to foreground or background by taking the inner product of the Fiedler vector of a filtered Gramian matrix over the tokens.
Contrary to \citet{wang2023tokencut}, we find that using the final output features yields better results for all models except MAE\@. 
Our reported results are thus for the out features for all models in our study, except MAE, for which we use the key features.
We set $\tau_\text{TC}=0.3$ for all models; \Cref{tab:tau-ablation} shows this setting yield better results for DINOv2 and CAPI\@.
Otherwise, we follow original implementation.

\paragraph{Selecting foreground partition using salient principal components.}
We also observe that some principal components have a strong center bias in the activations. This can be partially explained by object center bias in the training data. 
Comparing with the activation maps of synthetic input data in \cref{fig:activations_bigfig} shows that in several cases the center bias is not present.
This indicates that these components are responding to relative saliency or instance level correlations for each of the local patches, rather than an encoded positional center bias in the model.

Given a principal component $v_d$ with salient activations, we can effectively determine which patches are likely to be part of the foreground or class level object.
We find that we can improve zero-shot salient segmentation with TokenCut by selecting the foreground partition based on the patch with the highest response $\langle v_d, z \rangle$. In contrast, TokenCut selects the patch with maximum absolute value in its feature vector.
The results in \Cref{tab:salient_guide} show out-of-the-box improvement across the board, where we use the strongest salient principal component to guide foreground selection for each model.
We show results for all MIM models in our study, except I-JEPA which did not have a good salient principal component. We observed low performance on salient segmentation for all our experiments with I-JEPA, which suggests that the patch embeddings are not informative for this task.

\begin{table}[tb]
    \footnotesize
    \centering
    \caption{Ablation over the TokenCut parameter $\tau_\text{TC}$ on ECSSD~\citep{ecssd} for DINOv2 and CAPI.}
    \begin{tblr}{
        colspec={ccccccc},
        stretch=0,
    }
    \toprule
    & \SetCell[c=3]{c} DINOv2 ViT-B/16 &&& \SetCell[c=3]{c} CAPI ViT-L/14 && \\
    \cmidrule[r]{2-4} 
    \cmidrule[]{5-7}
    $\tau_\text{TC}$ & max $F_\beta$ & IoU & Acc. & max $F_\beta$ & IoU & Acc. \\
    \midrule
    0.2 & 79.037 & 70.033 & 87.720 & 64.927 & 53.906 & 78.027 \\
    0.3 & 79.803 & 71.751 & 87.953 & 72.456 & 66.083 & 84.334 \\
    0.4 & 79.177 & 71.708 & 87.655 & 70.604 & 64.627 & 84.139 \\
    \bottomrule
    \end{tblr}
    \label{tab:tau-ablation}
\end{table}

\begin{table}[tb]
    \footnotesize
    \centering
    \caption{Using salient principal components to select foreground partition in TokenCut. Results are shows for corrected embeddings after cleaning with \METHODNAME.}
    \begin{tblr}{
        colspec={llcccccc},
        colsep=2pt,
        stretch=0,
        cell{1}{6-8}={bg=highlight},
    }
    \toprule
    && \SetCell[c=3]{c} Max. abs. val. &&& \SetCell[c=3]{c} Max. Sal. PC reponse && \\
    \cmidrule[r]{3-5} 
    \cmidrule[]{6-8} 
    Model & Arch. & max $F_\beta$ & IoU & Acc. & max $F_\beta$ & IoU & Acc. \\
    \midrule
    DINOv2 & ViT-B/14 & 71.461 & 64.129 & 83.292 & 80.633 & 72.559 & 88.687 \\
    DINOv3 &  ViT-B/16 & 33.360 & 25.111 & 46.521 & 42.633 & 33.742 & 61.975 \\
    iBOT & ViT-B/16 & 64.432 & 57.978 & 80.093 & 66.557 & 60.167 & 78.340 \\
    CAPI & ViT-L/14 & 74.506 & 68.045 & 86.3234 & 85.219 & 78.084 & 92.600 \\
    MAE & ViT-B/16 & 80.525 & 70.705 & 90.506 & 82.094 & 72.118 & 91.444 \\
    Franca & ViT-B/14 & 77.660 & 70.830 & 87.577 & 84.176 & 76.985 & 91.514 \\
    \bottomrule
    \end{tblr}
    \label{tab:salient_guide}
\end{table}

\subsection{kNN segmentation}\label{sec:knnseg-details}

\begin{table*}[tb]
    \footnotesize
    \centering
    \caption{kNN segmentation comparison of original vs.\ \METHODNAME-corrected embeddings across backbones, reporting mean IoU and pixel accuracy (Acc) for the ADE20 \cite{zhou2017sceneADE20k, zhou2019semanticADE20k} and PascalVOC \cite{pascalvoc} benchmarks.}
    \begin{tblr}{
        colspec={llcccccccccccc},
        colsep=2pt,
        stretch=0,
        cell{2}{5-6}={bg=highlight},
        cell{2}{11-12}={bg=highlight},
        column{8,9}={colsep=4pt},
        column{6,8}={font=\tiny, fg=green!65!black, leftsep=0pt},
        column{12,14}={font=\tiny, fg=green!65!black, leftsep=0pt},
    }
    \toprule
    && \SetCell[c=6]{c} ADE20k &&&&&& \SetCell[c=6]{c} PascalVOC &&&&&& \\
    \cmidrule[lr]{3-8}
    \cmidrule[lr]{9-14}
    && \SetCell[c=2]{c} Original emb. && \SetCell[c=4]{c} Corrected emb. &&&&
    \SetCell[c=2]{c} Original emb. && \SetCell[c=4]{c} Corrected emb. & \\
    Pretrain & Model &
    {IoU} & {Acc}  & 
    \SetCell[c=2]{}IoU && \SetCell[c=2]{} Acc &&
    {IoU} & {Acc}  & 
    \SetCell[c=2]{}IoU && \SetCell[c=2]{} Acc & \\
    \midrule
    DINOv2 & ViT-B16 &
         40.25 & 74.60 & 
         40.81 & $\uparrow$\,0.56  & 74.72 & $\uparrow$\,0.12 &
         73.34 & 93.74 & 74.07 & $\uparrow$\ 0.73 & 93.90 & $\uparrow$\ 0.16 \\
    DINOv3 & ViT-B16 & 
         43.85 & 77.94 & 
         44.58 & $\uparrow$\,0.73  & 78.10 & $\uparrow$\,0.16 & 
         78.46 & 95.41 & 79.43 & $\uparrow$\, 0.97 & 95.58 & $\uparrow$\, 0.17 \\
    iBOT & ViT-B16 & 
         27.73 & 70.86 & 
         28.43 & $\uparrow$\,0.70  & 71.27 & $\uparrow$\,0.40 &
         61.10 & 91.25 & 62.16 & $\uparrow$\, 1.06 & 91.50 & $\uparrow$\, 0.25 \\
    CAPI & ViT-L14 & 
         31.38 & 71.63 & 
         31.64 & $\uparrow$\,0.26  & 71.77 & $\uparrow$\,0.14  &
         60.40 & 91.40 & 60.80 & $\uparrow$\, 0.40 & 91.50 & $\uparrow$\, 0.09\\
    MAE & ViT-B16 & 
         11.88 & 58.00 & 
         12.65 & $\uparrow$\,0.77  & 58.60 & $\uparrow$\,0.59 &
         27.99 & 82.624 & 30.39 & $\uparrow$\, 2.37 & 83.27 & $\uparrow$\, 0.65 \\
    I-JEPA & ViT-H14 & 
         20.95 & 60.27 & 
         21.26 & $\uparrow$\,0.31  & 60.29 & $\uparrow$\,0.01 &
         57.77 & 89.22 & 58.50 & $\uparrow$\, 0.73 & 89.34 & $\uparrow$\, 0.13 \\
    DINO & ViT-B16 &
         21.21 & 66.49 & 
         21.21 & \SetCell{fg=gray} \,0.00 & 66.49 & \SetCell{fg=gray} \,0.00 &
         47.32 & 88.01 & 47.32 & \SetCell{fg=gray} \, 0.00 & 88.01 & \SetCell{fg=gray} \, 0.00 \\
    \bottomrule
    \end{tblr}
    \label{tab:seg_additional}
\end{table*}

We evaluate segmentation quality without any learnable parameters using a patch-level $k$-nearest neighbor approach.
We first extract patch features from all training images using the frozen backbone and build a feature bank of $\ell_2$-normalized patch embeddings paired with their ground truth labels, obtained by downsampling the segmentation masks to the patch grid via nearest-neighbor interpolation.
At evaluation time, each query patch is classified by retrieving its $k=30$ nearest neighbors from the feature bank using cosine similarity, with temperature-scaled ($\tau=0.07$) weighted voting over the neighbor labels.
This approach directly probes the spatial quality of frozen patch representations without introducing any learnable parameters, making it a good diagnostic for raw feature quality.

In \cref{tab:seg_additional} we provide additional results for PascalVOC \cite{pascalvoc}, expanding the results for ADE20k~\citep{zhou2017sceneADE20k, zhou2019semanticADE20k} that were reported in \cref{tab:seg} in the main paper.

\subsection{kNN classification}\label{sec:knncls-details}

In this section we provide details on our kNN classification evaluation protocol.
In the main article we report kNN classification by attention- and entropy-weighted aggregation of patch predictions; the exact details are in \cref{sec:cls-weighted-details}. Furthermore, we report kNN classification on the averaged patch embeddings in \cref{sec:cls-avg-details} for completeness.

\subsubsection{kNN classification by weighted aggregation of patch predictions}
\label{sec:cls-weighted-details}

\begin{table*}[tb]
    \footnotesize
    \centering
    \caption{Weighted kNN classification by aggregating patch prediction on iNat2018 \cite{inat2018} and ImageNet \cite{imagenet}, expanding results from \cref{tab:knncls-weighted}.
    We compare original vs. \METHODNAME-corrected embeddings across backbones.
    }
    \begin{tblr}{
        colspec={llcccccccccccc},
        colsep=2pt,
        stretch=0,
        cell{2}{5-6}={bg=highlight},
        cell{2}{11-12}={bg=highlight},
        column{8,9}={colsep=4pt},
        column{6,8}={font=\tiny, fg=green!65!black, leftsep=0pt},
        column{12,14}={font=\tiny, fg=green!65!black, leftsep=0pt},
    }
    \toprule
    && \SetCell[c=6]{c} ImageNet &&&&&& \SetCell[c=6]{c} iNat &&&&&& \\
    \cmidrule[lr]{3-8}
    \cmidrule[lr]{9-14}
    && \SetCell[c=2]{c} Original emb. && \SetCell[c=4]{c} Corrected emb. &&&&
    \SetCell[c=2]{c} Original emb. && \SetCell[c=4]{c} Corrected emb. & \\
    Pretrain & Model &
    {Acc@1} & {Acc@5}  & 
    \SetCell[c=2]{}Acc@1 && \SetCell[c=2]{} Acc@5 &&
    {Acc@1} & {Acc@5}  & 
    \SetCell[c=2]{}Acc@1 && \SetCell[c=2]{} Acc@5 && \\
    \midrule
    DINOv2 & ViT-B16 & 
         82.32 & 96.29 & 
         82.59 & $\uparrow$\,0.27  & 96.30 & $\uparrow$\,0.01 &
         58.80 & 82.99 & 
         60.36 & $\uparrow$\ 1.56 & 83.75 & $\uparrow$\ 0.76 \\
    DINOv3 & ViT-B16 & 
         81.47 & 95.57 & 
         81.48 & $\uparrow$\,0.01  & 95.59 & $\uparrow$\,0.01 &
         58.25 & 79.52 & 
         58.76 & $\uparrow$\ 0.51 & 80.12 & $\uparrow$\ 0.60 \\
    iBOT & ViT-B16 & 
         71.45 & 90.03 & 
         71.60 & $\uparrow$\,0.15  & 90.10 & $\uparrow$\,0.07 &
         27.64  & 49.06 & 
         27.70 & $\uparrow$\ 0.06 & 49.73 & $\uparrow$\ 0.66 \\
    CAPI\smash{\textsuperscript{\dag}} & ViT-L14 & 
         70.81 & 91.09 & 
         71.25 & $\uparrow$\,0.43  & 91.33 & $\uparrow$\,0.24 &
         26.88 & 51.18 & 
         27.64 & $\uparrow$\ 0.76 & 52.25 & $\uparrow$\ 1.07 \\
    MAE\smash{\textsuperscript{\dag}} & ViT-B16 & 
         59.30 & 81.49 & 
         60.62 & $\uparrow$\,1.32  & 82.38 & $\uparrow$\,0.89  &
         16.43 & 31.36 & 
         19.12 & $\uparrow$\ 2.70 & 36.61 & $\uparrow$\ 5.25 \\
    I-JEPA\smash{\textsuperscript{\dag}} & ViT-H14 & 
         75.68 & 91.70 & 
         75.89 & $\uparrow$\,0.21  & 91.76 & $\uparrow$\,0.06 &
         17.29 & 35.34 &  
         18.01 & $\uparrow$\ 0.73 & 36.06 & $\uparrow$\ 0.72 \\
    DINO & ViT-B16 &
         66.08 & 86.13 & 
         66.08 & \SetCell{fg=gray} \,0.00  & 86.13 & \SetCell{fg=gray} \,0.00 &
         24.64 & 44.60 & 
         24.64 & \SetCell{fg=gray} \, 0.00 & 44.60 & \SetCell{fg=gray} \, 0.00 \\
    \bottomrule
    \end{tblr}
    \\[1pt]
    {\scriptsize \textsuperscript{\dag}Aggregation weighted by entropy for models with no class token objective; otherwise weighted by class attention.}
    \label{tab:knncls-weighted-additional}
\end{table*}

In \cref{tab:knncls-weighted} in \cref{sec:downstream-performance}, we evaluate frozen patch features on ImageNet~\citep{imagenet} using a patch-level $k$-nearest neighbor classification protocol with $k=20$ and temperature $\tau=0.07$.
We extract patch tokens from the last layer of each frozen backbone and reduce their dimensionality to $256$ using PCA.
Features are $\ell_2$-normalized and stored in a feature bank sharded across $2$ GPUs in float16 precision.
Each patch in a query image retrieves its $k$ nearest neighbors from the training feature bank via cosine similarity, producing per-patch class probability distributions through temperature-scaled softmax weighting.
The per-patch probabilities are then aggregated into a single image-level prediction using one of two weighting schemes:

\begin{enumerate}
    \item \textit{CLS attention weighting}, which uses the attention weights from the \texttt{[CLS]} token in the last self-attention layer of the backbone to weight each patch's contribution, thus leveraging the model's own learned notion of patch importance.
    \item \textit{Entropy weighting}, which assigns higher weight to patches whose $k$-NN probability distributions have lower entropy, favoring patches that yield more confident predictions.
\end{enumerate}

We aggregate by CLS attention weighting for models trained with an instance discrimination objective (DINO, DINOv2, DINOv3, iBOT), as these methods explicitly train the \texttt{[CLS]} token to capture global image semantics through their contrastive or self-distillation losses, yielding meaningful attention distributions over patches.
For models trained exclusively with a masked image modeling objective (I-JEPA, MAE, CAPI), the \texttt{[CLS]} token is either absent or not trained to aggregate global information, so its attention weights are not informative for patch weighting.
We therefore use entropy-based aggregation for these models, which is agnostic to the pretraining objective and instead relies on the confidence of the per-patch $k$-NN predictions themselves.

In \cref{tab:knncls-weighted-additional} we provide additional results for iNat2018 \cite{inat2018}, expanding the results for ImageNet \cite{imagenet} that were reported in \cref{tab:knncls-weighted} in the main paper.

\subsubsection{kNN classification on averaged patch embeddings.}
\label{sec:cls-avg-details}

We additionally perform kNN classification on  ImageNet~\citep{imagenet} by average pooling the patch embeddings, and matching the validation embeddings to the $k$ nearest embeddings from the training set.
We follow the kNN evaluation script by \citet{caron2021emerging}, and set $k=20$ number of neighbors and $0.07$ temperature for the voting coefficient.

We compare the top-1 and top-5 accuracies of the average of the patch embeddings, and the corrected patch embeddings.
The results in \Cref{tab:knncls} show only modest improvements after correcting for invariant components.
This is the expected result, as instance tasks are not as reliant on local semantics, and the benefit of \METHODNAME may be dampened by the uniform aggregation, as spatial information about the locality of patches reduces due to averaging them out.

\begin{table}[tb]
    \footnotesize
    \centering
    \caption{kNN classification of average pooled patch embeddings on ImageNet~\citep{imagenet}. We compare the original embeddings with the \METHODNAME-corrected embeddings for each backbone, and report top-1 and top-5 accuracies.}
    \begin{tblr}{
        colspec={llcccccc},
        colsep=2pt,
        stretch=0,
        cell{1}{5-6}={bg=highlight},
        column{6,8}={font=\tiny, fg=green!65!black, leftsep=0pt},
    }
    \toprule
    & & \SetCell[c=2]{c} Original embeddings && \SetCell[c=4]{c} Corrected embeddings & \\
    \cmidrule[lr]{3-4} 
    \cmidrule[lr]{5-8} 
    Pretrain & Model &
    {Acc@1} & {Acc@5}  & 
    \SetCell[c=2]{}Acc@1 && \SetCell[c=2]{} Acc@5 & \\
    \midrule
    DINOv2 & ViT-B16 & 
        77.064 & 91.624 & 
        77.100 & $\uparrow$\, 0.036 & 91.636 & $\uparrow$\, 0.012 \\
    DINOv3 & ViT-B16 & 
        76.542 & 91.530 & 
        76.588 & $\uparrow$\, 0.046 & 91.612 & $\uparrow$\, 0.082 \\
    iBOT & ViT-B16 & 
        59.170 & 79.612 & 
        59.498 & $\uparrow$\, 0.328 & 79.918 & $\uparrow$\, 0.306 \\
    CAPI & ViT-L14 & 
        56.250 & 77.490 & 
        56.444 & $\uparrow$\, 0.194 & 77.742 & $\uparrow$\, 0.252 \\
    MAE & ViT-B16 & 
        47.488 & 69.168 & 
        47.758 & $\uparrow$\, 0.27 & 69.442 & $\uparrow$\, 0.274 \\
    I-JEPA & ViT-H14 & 
        71.382 & 86.144 & 
        71.390 & $\uparrow$\, 0.008 & 86.168 & $\uparrow$\, 0.024 \\
    DINO & ViT-B16 &
        55.216 & 75.740 &
        55.216 & \SetCell{fg=gray} \, 0.000 & 75.740 & \SetCell{fg=gray} \, 0.000 \\
    \bottomrule
    \end{tblr}
    \label{tab:knncls}
\end{table}

\section{Linear evaluation protocols}\label{sec:lin-eval}

Our intention with restricting evaluation to zero-shot protocols (TokenCut, kNN) is to measure the \textit{intrinsic fidelity} of the representations, while \textit{limiting confounding factors}.
As \METHODNAME uses linear PCA, a learnable head can adapt to suppress positional noise when this information is unhelpful for the task.
Linear evaluation protocols are thus unsuitable for diagnosing positional noise, as they can adapt, unintentionally masking the issue.
Indeed, linear probing, attentive probing, and linear segmentation yield similar results with \METHODNAME; 
see \cref{tab:lincls}, \cref{tab:attcls}, and \cref{tab:linseg}.
The difference in performance with and without \METHODNAME is very low---this level of variation in performance is expected when probing with learnable heads,
and the difference is thus too small to attribute any change in performance to \METHODNAME.
We describe each of these evaluation protocols in detail below.

In contrast, kNN directly probes representation geometry without learned transformations, making it a faithful diagnostic for raw feature quality~\citep{wu2018, caron2021emerging}.
This matters in practice: using SSL representations out-of-the-box is common, and practitioners unaware of positional noise may encounter false positives from patch embeddings at similar relative locations.

\subsection{Linear evaluation.} We perform linear evaluation on the averaged patch embeddings of the last output layer of each model on ImageNet~\citep{imagenet}. 
We follow the standard protocol of training a single linear layer for classification on top of the frozen features for $100$ epochs.
We use a standard stochastic gradient descent optimizer (\texttt{SGD}) with a base learning rate of $0.001$, momentum $0.9$, cosine learning rate decay, and a batch size of $256$ with $4$ GPUs (effective batch size $1024$).
Following protocol, the learning rate is scaled by 
\begin{equation*}
    \text{lr} = \frac{\text{base lr} \times \text{batch size} \times \text{num GPUs}}{256}.
\end{equation*}
The results in \cref{tab:lincls} show minor changes in performance when the embeddings are corrected with \METHODNAME, reflecting that suppressing positional noise has little effect when the classification head is learnable.

\subsection{Attentive probing.}
The attentive probing protocol replaces the global average pooling of patch tokens with a learnable attention mechanism that computes a weighted aggregation over the patch tokens before classification~\citep{bardes2024revisiting}.
Specifically, a lightweight two-layer MLP computes per-patch attention logits, which are normalized via softmax to produce attention weights over the spatial positions.
The attended feature is then passed to a linear classifier.
This allows the probe to selectively focus on the most informative patches, which is particularly beneficial for methods where discriminative information is distributed across patch tokens rather than concentrated in a single global representation.
We otherwise follow the linear evaluation protocol, training for $100$ epochs with \texttt{SGD}, a base learning rate of $0.0025$, momentum $0.9$, cosine learning rate decay, and a batch size of $256$ with $4$ GPUs (effective batch size $1024$).
The results in \cref{tab:attcls} show minor changes in performance when the embeddings are corrected with \METHODNAME, reflecting that suppressing positional noise has little effect when the classification head is learnable.

\subsection{Linear Segmentation.}
We evaluate segmentation quality by training a linear segmentation head on top of frozen patch features on ADE20k~\citep{zhou2017sceneADE20k, zhou2019semanticADE20k}.
The segmentation head consists of a single $1\times1$ convolution applied to the spatial patch feature map, followed by bilinear upsampling to the original image resolution.
We train with cross-entropy loss and \texttt{SGD} with momentum $0.9$, polynomial learning rate decay with power $0.9$, and a batch size of $32$ per GPU across $4$ GPUs (effective batch size $128$) for $80$ epochs.
The learning rate is selected from $\{0.08, 0.04, 0.008\}$ based on validation mIoU for the baseline results of the original embeddings for each model.
The crop size is set to be divisible by the patch size of the backbone: $512$ for patch size $16$ and $518$ for patch size $14$.
Training images are augmented with random scaling (ratio $0.5$--$2.0\times$), random cropping, and random horizontal flipping; validation images are resized to the crop size.
We report mean intersection over union (IoU) and per pixel accuracy in \cref{tab:linseg}, once again showing that learnable evaluation heads confound the effect of suppressing positional noise with \METHODNAME.

\begin{table}[tb]
    \footnotesize
    \centering
    \caption{Linear evaluation protocol.}
    \begin{tblr}{
        colspec={llcccccc},
        colsep=2pt,
        stretch=0,
        cell{1}{5-6}={bg=highlight},
        column{6,8}={font=\tiny, fg=green!65!black, leftsep=0pt},
    }
    \toprule
    & & \SetCell[c=2]{c} Original embeddings && \SetCell[c=4]{c} Corrected embeddings & \\
    \cmidrule[lr]{3-4} 
    \cmidrule[lr]{5-8} 
    Pretrain & Model &
    {Acc@1} & {Acc@5}  & 
    \SetCell[c=2]{}Acc@1 && \SetCell[c=2]{} Acc@5 & \\
    \midrule
    DINOv2 & ViT-B16 & 
        81.13 & 96.01 & 
        81.15 & $\uparrow$\,0.02 & 96.01 & \SetCell{fg=gray} 0.00 \\
    DINOv3 & ViT-B16 & 
        76.77 & 93.89 & 
        76.76 & \SetCell{fg=red!65!black} $\downarrow$\,-0.01 & 93.90 & $\uparrow$\, 0.01 \\
    iBOT & ViT-B16 & 
        72.89 & 91.37 & 
        72.93 & $\uparrow$\,0.04 & 91.35 & \SetCell{fg=red!65!black} $\downarrow$\, -0.02 \\
    CAPI & ViT-L14 & 
        63.09 & 85.41 & 
        63.17 & $\uparrow$\,0.08 & 85.46 & $\uparrow$\, 0.05 \\
    MAE & ViT-B16 & 
        50.60 & 74.83 & 
        50.63 & $\uparrow$\,0.03 & 74.87 & $\uparrow$\, 0.04 \\
    I-JEPA & ViT-H14 & 
        74.99 & 90.65 & 
        74.99 & \SetCell{fg=gray} 0.00 & 90.69 & $\uparrow$\, 0.04 \\
    DINO & ViT-B16 &
        66.36 & 86.43 & 
        66.39 & $\uparrow$\,0.03 & 86.41 & \SetCell{fg=red!65!black} $\downarrow$\, -0.02 \\
    \bottomrule
    \end{tblr}
    \label{tab:lincls}
\end{table}

\begin{table}[tb]
    \footnotesize
    \centering
    \caption{Attentive probing.}
    \begin{tblr}{
        colspec={llcccccc},
        colsep=2pt,
        stretch=0,
        cell{1}{5-6}={bg=highlight},
        column{6,8}={font=\tiny, fg=green!65!black, leftsep=0pt},
    }
    \toprule
    & & \SetCell[c=2]{c} Original embeddings && \SetCell[c=4]{c} Corrected embeddings & \\
    \cmidrule[lr]{3-4} 
    \cmidrule[lr]{5-8} 
    Pretrain & Model &
    {Acc@1} & {Acc@5}  & 
    \SetCell[c=2]{}Acc@1 && \SetCell[c=2]{} Acc@5 & \\
    \midrule
    DINOv2 & ViT-B16 & 
        84.97 & 97.20 & 
        85.00 & $\uparrow$\,0.03 & 97.22 & $\uparrow$\, 0.02 \\
    DINOv3 & ViT-B16 & 
        83.40 & 96.56 & 
        83.46 & $\uparrow$\,0.06 & 96.61 & $\uparrow$\, 0.05 \\
    iBOT & ViT-B16 & 
        79.05 & 94.34 & 
        79.04 & \SetCell{fg=red!65!black} $\downarrow$\,-0.01 & 94.31 & \SetCell{fg=red!65!black} $\downarrow$\, -0.03 \\
    CAPI & ViT-L14 & 
        81.75 & 95.84 & 
        81.72 & \SetCell{fg=red!65!black} $\downarrow$\,-0.03 & 95.93 & $\uparrow$\, 0.09 \\
    MAE & ViT-B16 & 
        67.80 & 87.44 & 
        67.90 & $\uparrow$\,0.10 & 87.47 & $\uparrow$\, 0.03 \\
    I-JEPA & ViT-H14 & 
        77.66 & 92.84 & 
        77.60 & \SetCell{fg=red!65!black} $\downarrow$\,-0.06 & 92.77 & \SetCell{fg=red!65!black} $\downarrow$\, -0.07 \\
    DINO & ViT-B16 &
        72.56 & 90.58 & 
        72.50 & \SetCell{fg=red!65!black} $\downarrow$\,-0.06 & 90.58 & \SetCell{fg=gray} 0.00 \\
    \bottomrule
    \end{tblr}
    \label{tab:attcls}
\end{table}

\begin{table}[tb]
    \footnotesize
    \centering
    \caption{Linear segmentation.}
    \begin{tblr}{
        colspec={llcccccc},
        colsep=2pt,
        stretch=0,
        cell{1}{5-6}={bg=highlight},
        column{6,8}={font=\tiny, fg=green!65!black, leftsep=0pt},
    }
    \toprule
    & & \SetCell[c=2]{c} Original embeddings && \SetCell[c=4]{c} Corrected embeddings & \\
    \cmidrule[lr]{3-4} 
    \cmidrule[lr]{5-8} 
    Pretrain & Model &
    {IoU} & {Acc}  & 
    \SetCell[c=2]{}IoU && \SetCell[c=2]{} Acc & \\
    \midrule
    DINOv2 & ViT-B16 & 
        47.54 & 80.21 & 
        47.53 & \SetCell{fg=red!65!black} $\downarrow$\,-0.01 & 80.20 & \SetCell{fg=red!65!black} $\downarrow$\, -0.01 \\
    DINOv3 & ViT-B16 & 
        49.35 & 82.31 & 
        49.28 & \SetCell{fg=red!65!black} $\downarrow$\,-0.07 & 82.25 &\SetCell{fg=red!65!black} $\downarrow$\, -0.06 \\
    iBOT & ViT-B16 & 
        35.61 & 75.97 & 
        35.61 & $\uparrow$\,0.00 & 76.12 & $\uparrow$\, 0.15 \\
    CAPI & ViT-L14 & 
        41.96 & 78.73 & 
        41.98 & $\uparrow$\,0.02 & 78.55 & \SetCell{fg=red!65!black} $\downarrow$\, -0.18 \\
    MAE & ViT-B16 & 
        20.43 & 65.10 & 
        20.52 & $\uparrow$\,0.09 & 65.48 & $\uparrow$\, 0.38 \\
    I-JEPA & ViT-H14 & 
        27.43 & 68.71 & 
        27.53 & $\uparrow$\,0.10 & 68.77 & $\uparrow$\, 0.06 \\
    DINO & ViT-B16 &
        28.95 & 71.28 & 
        28.88 & \SetCell{fg=red!65!black} $\downarrow$\, -0.07 & 71.19 & \SetCell{fg=red!65!black} $\downarrow$\, -0.09 \\
    \bottomrule
    \end{tblr}
    \label{tab:linseg}
\end{table}

\section{Additional salient segmentation examples}

\begin{figure*}[tb]
    \centering
    \setlength{\tabcolsep}{2pt}
    \footnotesize
    \begin{tabular}{cccccc}
        Input & Raw pred. & \METHODNAME\ pred. & Raw pred. + BS & \METHODNAME pred. + BS & Ground Truth \\
        
        \includegraphics[width=0.155\linewidth, trim={30 50 50 50}, clip]{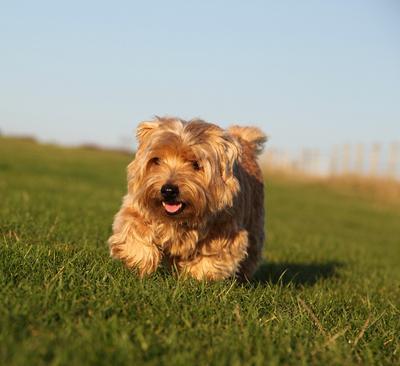} & 
        \includegraphics[width=0.155\linewidth, trim={30 50 50 50}, clip]{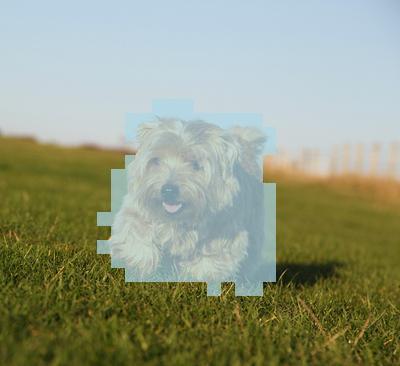} & 
        \includegraphics[width=0.155\linewidth, trim={30 50 50 50}, clip]{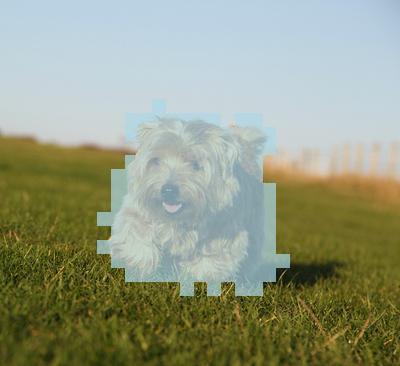} & 
        \includegraphics[width=0.155\linewidth, trim={30 50 50 50}, clip]{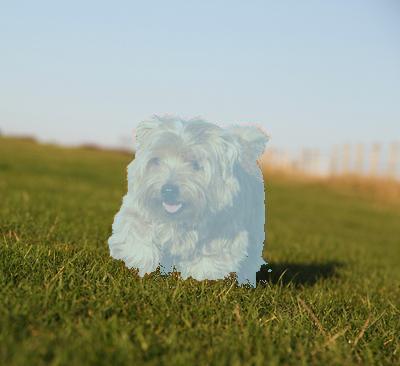} & 
        \includegraphics[width=0.155\linewidth, trim={30 50 50 50}, clip]{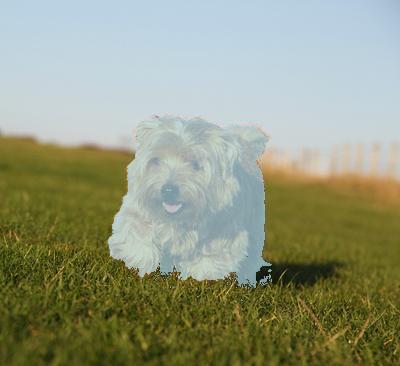} & 
        \includegraphics[width=0.155\linewidth, trim={30 50 50 50}, clip]{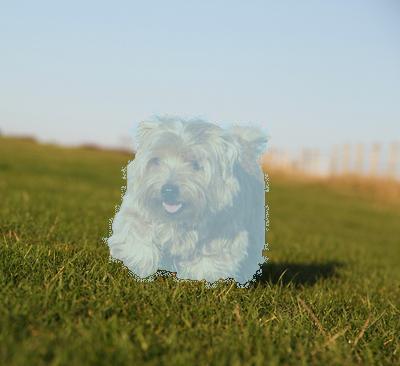} \\

        \includegraphics[width=0.155\linewidth, trim={30 0 50 0}, clip]{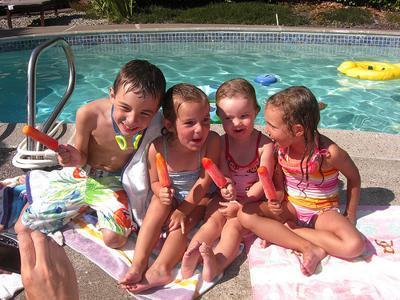} & 
        \includegraphics[width=0.155\linewidth, trim={30 0 50 0}, clip]{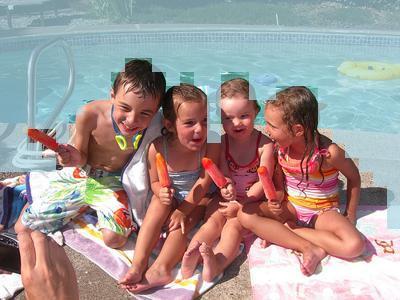} & 
        \includegraphics[width=0.155\linewidth, trim={30 0 50 0}, clip]{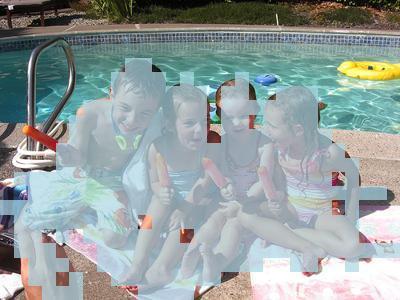} & 
        \includegraphics[width=0.155\linewidth, trim={30 0 50 0}, clip]{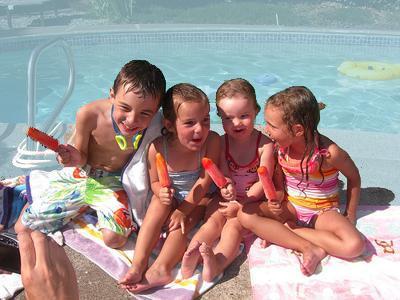} & 
        \includegraphics[width=0.155\linewidth, trim={30 0 50 0}, clip]{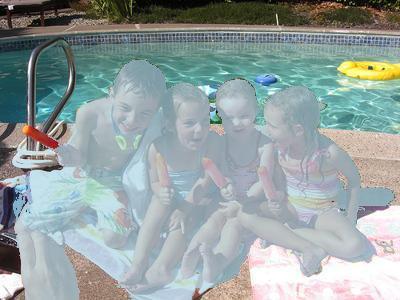} & 
        \includegraphics[width=0.155\linewidth, trim={30 0 50 0}, clip]{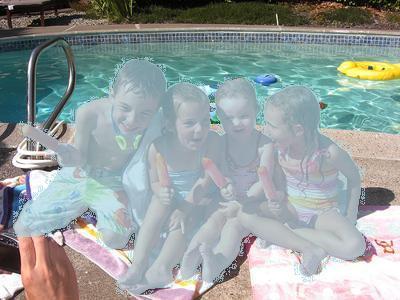} \\

        \includegraphics[width=0.155\linewidth, trim={50 15 50 15}, clip]{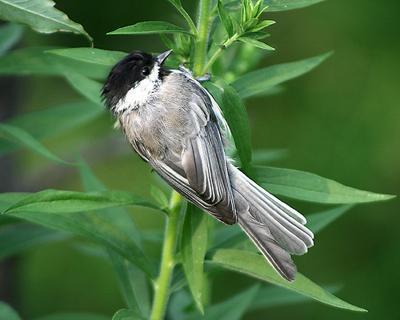} & 
        \includegraphics[width=0.155\linewidth, trim={50 15 50 15}, clip]{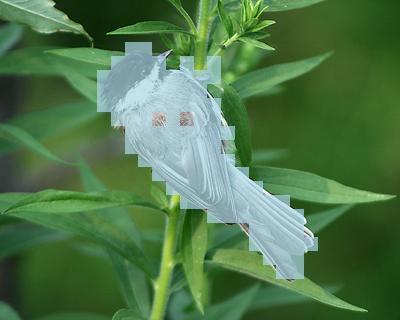} & 
        \includegraphics[width=0.155\linewidth, trim={50 15 50 15}, clip]{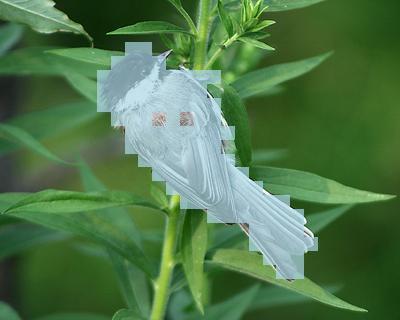} & 
        \includegraphics[width=0.155\linewidth, trim={50 15 50 15}, clip]{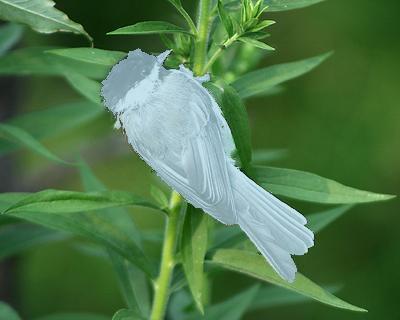} & 
        \includegraphics[width=0.155\linewidth, trim={50 15 50 15}, clip]{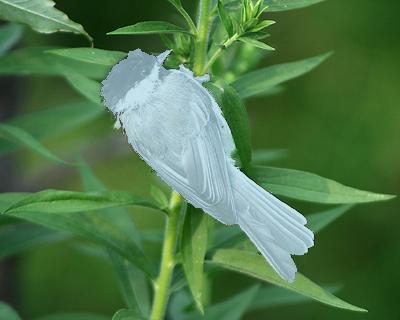} & 
        \includegraphics[width=0.155\linewidth, trim={50 15 50 15}, clip]{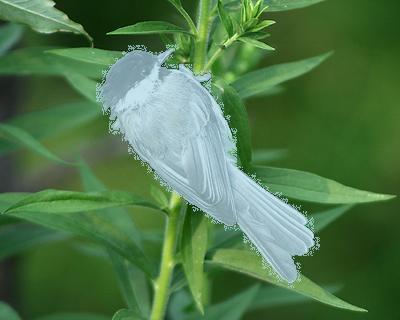} \\

        \includegraphics[width=0.155\linewidth, trim={0 100 50 0}, clip]{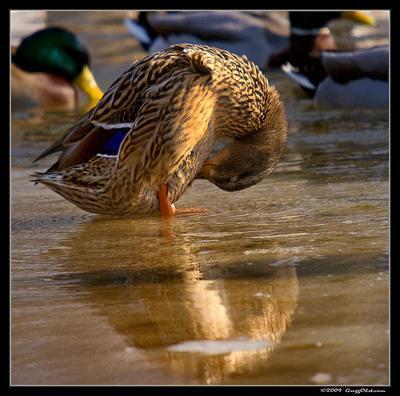} & 
        \includegraphics[width=0.155\linewidth, trim={0 100 50 0}, clip]{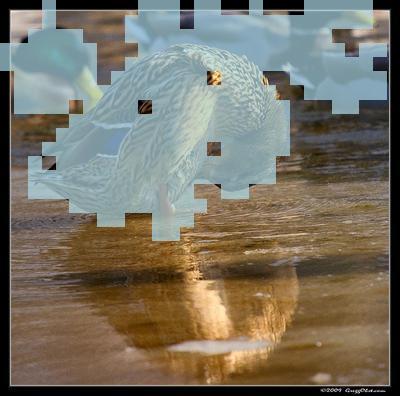} & 
        \includegraphics[width=0.155\linewidth, trim={0 100 50 0}, clip]{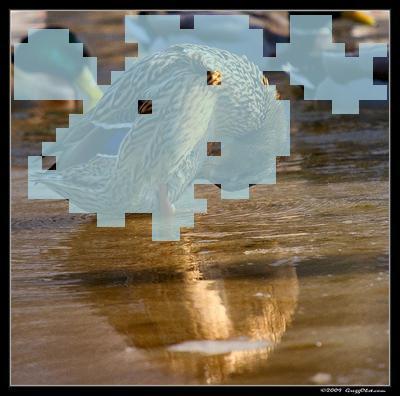} & 
        \includegraphics[width=0.155\linewidth, trim={0 100 50 0}, clip]{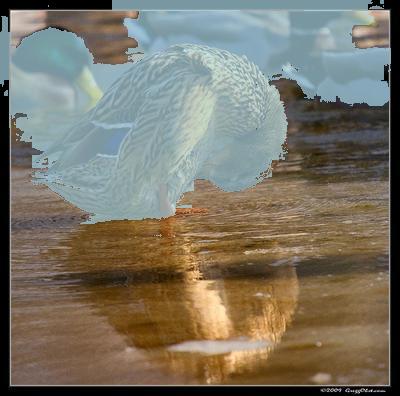} & 
        \includegraphics[width=0.155\linewidth, trim={0 100 50 0}, clip]{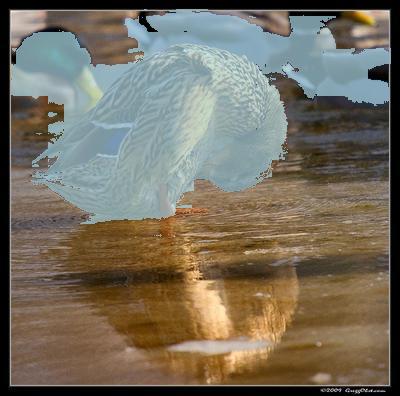} & 
        \includegraphics[width=0.155\linewidth, trim={0 100 50 0}, clip]{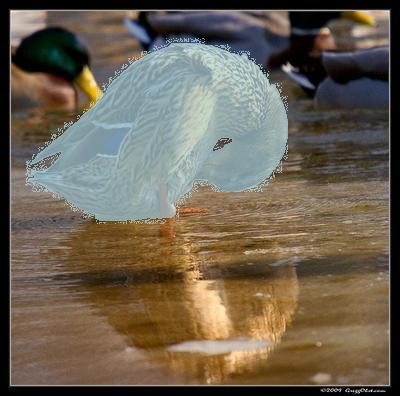} \\

        \includegraphics[width=0.155\linewidth, trim={0 10 0 50}, clip]{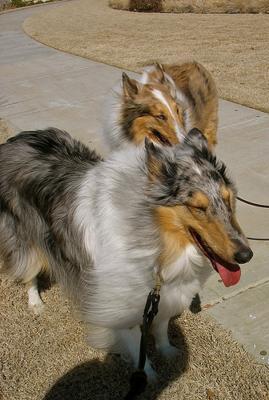} & 
        \includegraphics[width=0.155\linewidth, trim={0 10 0 50}, clip]{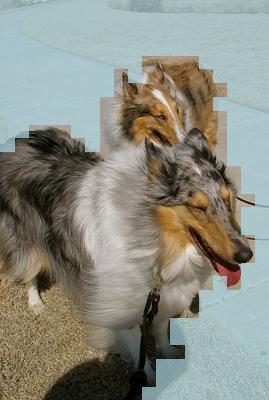} & 
        \includegraphics[width=0.155\linewidth, trim={0 10 0 50}, clip]{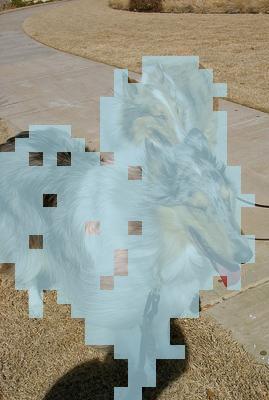} & 
        \includegraphics[width=0.155\linewidth, trim={0 10 0 50}, clip]{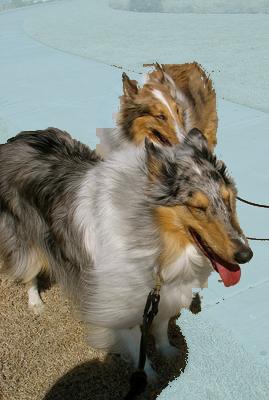} & 
        \includegraphics[width=0.155\linewidth, trim={0 10 0 50}, clip]{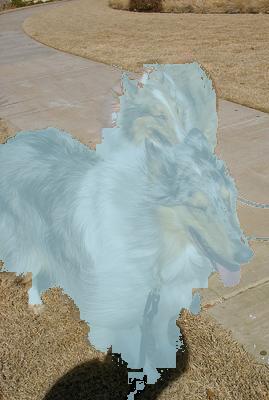} & 
        \includegraphics[width=0.155\linewidth, trim={0 10 0 50}, clip]{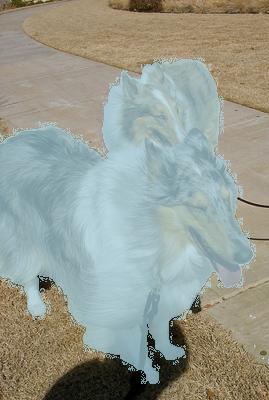} \\

        \includegraphics[width=0.155\linewidth, trim={50 0 50 0}, clip]{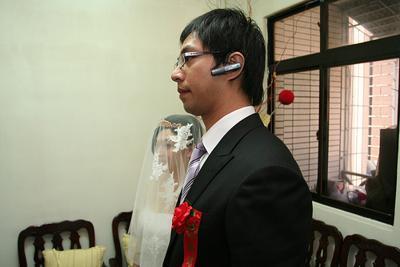} & 
        \includegraphics[width=0.155\linewidth, trim={50 0 50 0}, clip]{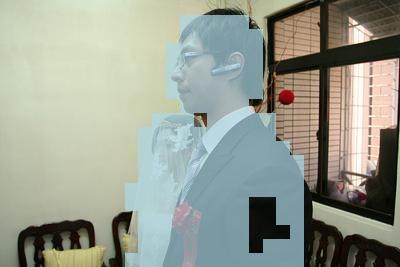} & 
        \includegraphics[width=0.155\linewidth, trim={50 0 50 0}, clip]{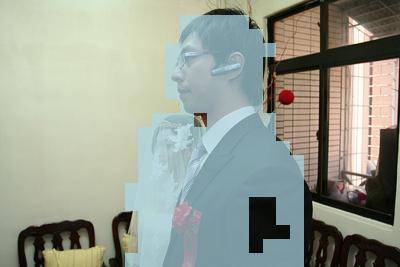} & 
        \includegraphics[width=0.155\linewidth, trim={50 0 50 0}, clip]{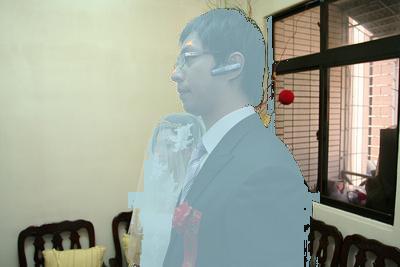} & 
        \includegraphics[width=0.155\linewidth, trim={50 0 50 0}, clip]{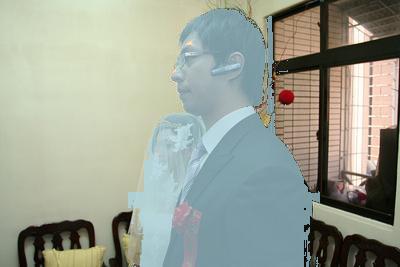} & 
        \includegraphics[width=0.155\linewidth, trim={50 0 50 0}, clip]{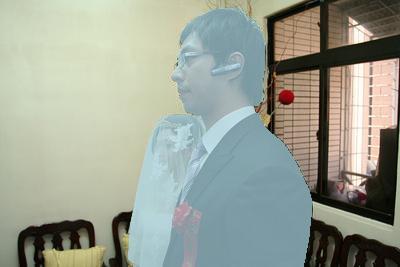} \\

        \includegraphics[width=0.155\linewidth, trim={0 0 0 0}, clip]{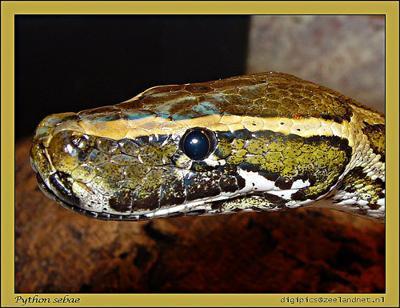} & 
        \includegraphics[width=0.155\linewidth, trim={0 0 0 0}, clip]{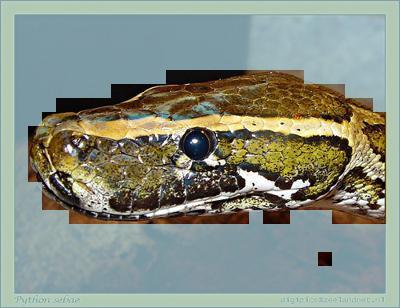} & 
        \includegraphics[width=0.155\linewidth, trim={0 0 0 0}, clip]{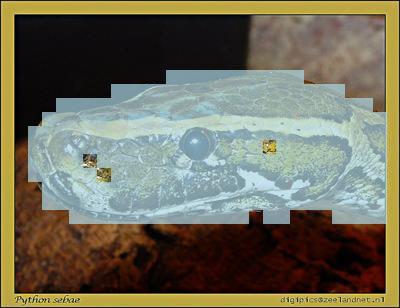} & 
        \includegraphics[width=0.155\linewidth, trim={0 0 0 0}, clip]{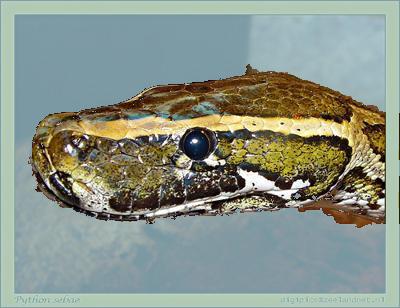} & 
        \includegraphics[width=0.155\linewidth, trim={0 0 0 0}, clip]{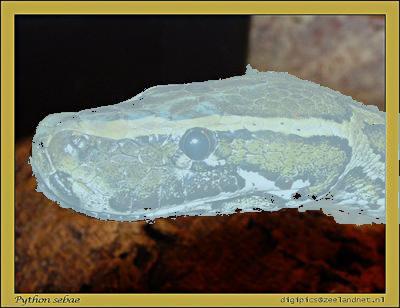} & 
        \includegraphics[width=0.155\linewidth, trim={0 0 0 0}, clip]{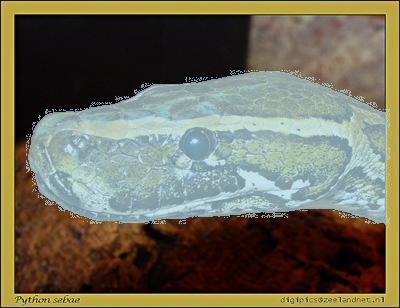} \\
        
    \end{tabular}
    \caption{Examples of salient segmentation from DUTS~\citep{duts} using TokenCut~\citep{wang2023tokencut} with frozen CAPI~\citep{darcet2025cluster} on the raw embeddings (Raw pred.) and after correcting with \METHODNAME (\METHODNAME pred.). 
    We show the predictions per patch and after refining the segmentation maps with the bilateral solver (BS).  
    Suppressing positional noise with \METHODNAME either matches or improves the zero-shot saliency maps.
    These examples are from the first samples in the dataset.
    }
    \label{fig:salient_improvement_examples_duts}
\end{figure*}

\begin{figure*}[tb]
    \centering
    \setlength{\tabcolsep}{2pt}
    \footnotesize
    \begin{tabular}{cccccc}
        Input & Raw pred. & \METHODNAME\ pred. & Raw pred. + BS & \METHODNAME pred. + BS & Ground Truth \\
        \includegraphics[width=0.155\linewidth, trim={10 0 10 100}, clip]{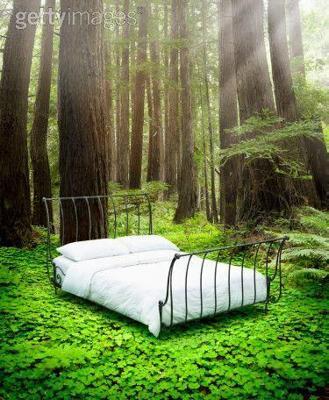} & 
        \includegraphics[width=0.155\linewidth, trim={10 0 10 100}, clip]{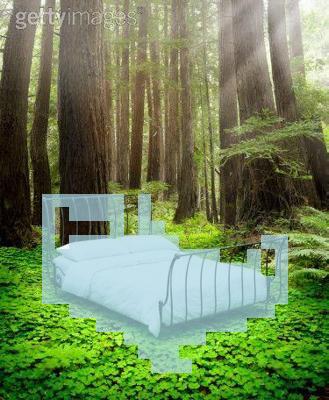} & 
        \includegraphics[width=0.155\linewidth, trim={10 0 10 100}, clip]{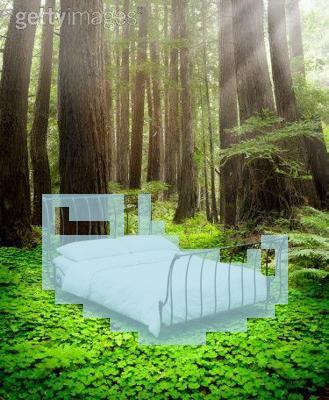} & 
        \includegraphics[width=0.155\linewidth, trim={10 0 10 100}, clip]{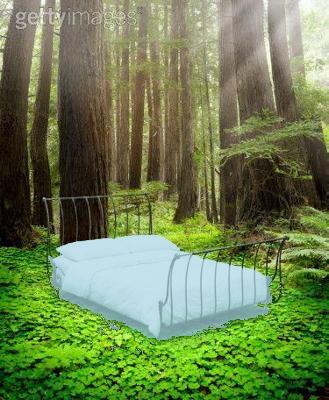} & 
        \includegraphics[width=0.155\linewidth, trim={10 0 10 100}, clip]{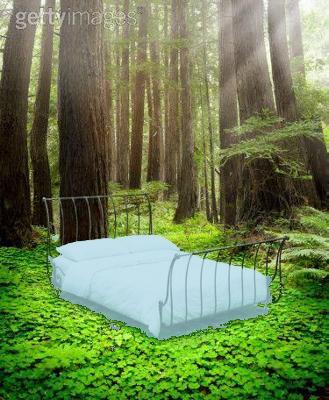} & 
        \includegraphics[width=0.155\linewidth, trim={10 0 10 100}, clip]{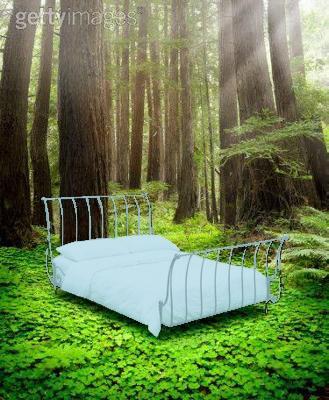} \\
        
        \includegraphics[width=0.155\linewidth, trim={50 0 50 0}, clip]{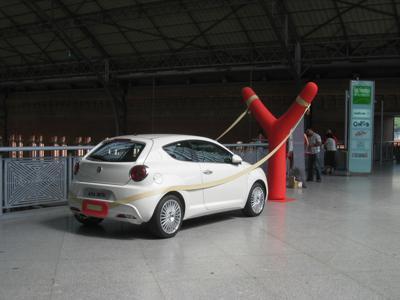} & 
        \includegraphics[width=0.155\linewidth, trim={50 0 50 0}, clip]{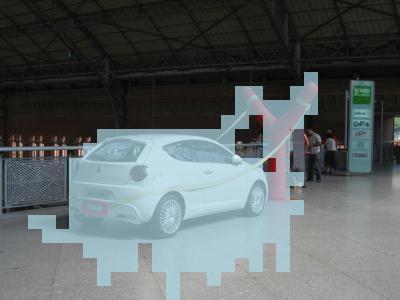} & 
        \includegraphics[width=0.155\linewidth, trim={50 0 50 0}, clip]{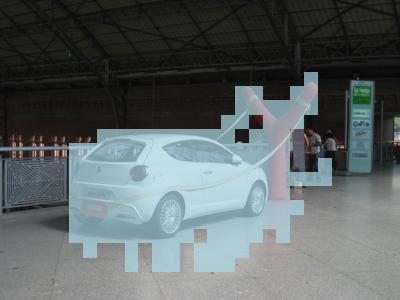} & 
        \includegraphics[width=0.155\linewidth, trim={50 0 50 0}, clip]{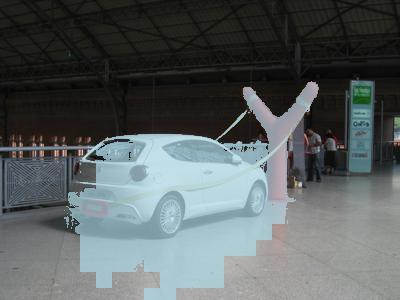} & 
        \includegraphics[width=0.155\linewidth, trim={50 0 50 0}, clip]{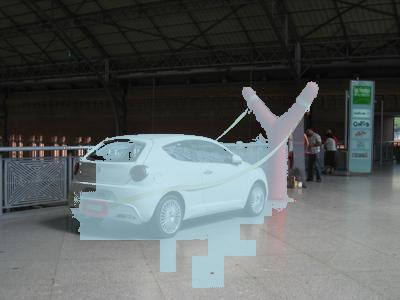} & 
        \includegraphics[width=0.155\linewidth, trim={50 0 50 0}, clip]{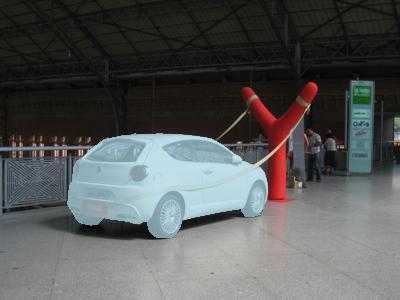} \\

        \includegraphics[width=0.155\linewidth, trim={50 20 50 100}, clip]{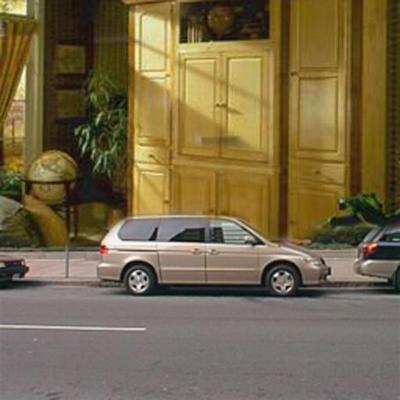} & 
        \includegraphics[width=0.155\linewidth, trim={50 20 50 100}, clip]{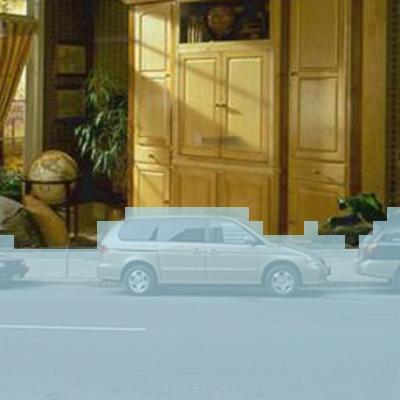} & 
        \includegraphics[width=0.155\linewidth, trim={50 20 50 100}, clip]{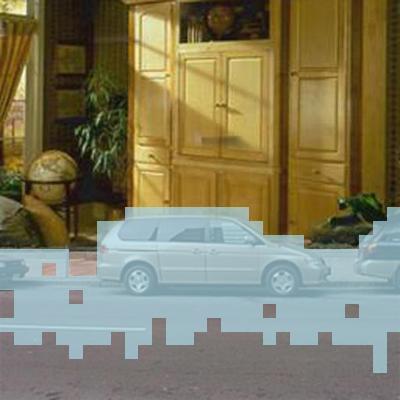} & 
        \includegraphics[width=0.155\linewidth, trim={50 20 50 100}, clip]{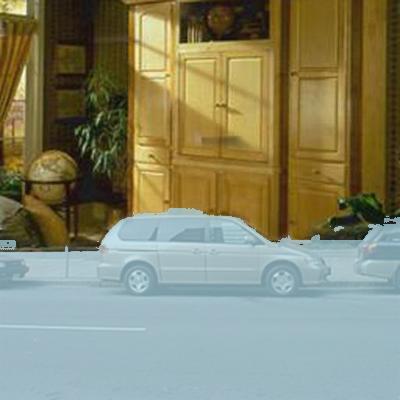} & 
        \includegraphics[width=0.155\linewidth, trim={50 20 50 100}, clip]{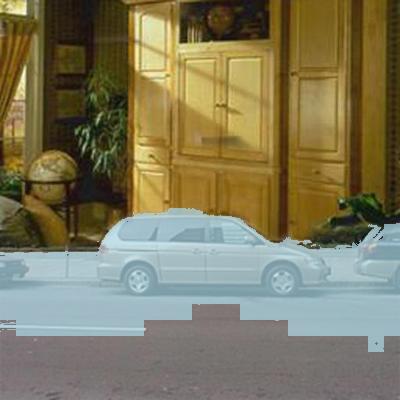} & 
        \includegraphics[width=0.155\linewidth, trim={50 20 50 100}, clip]{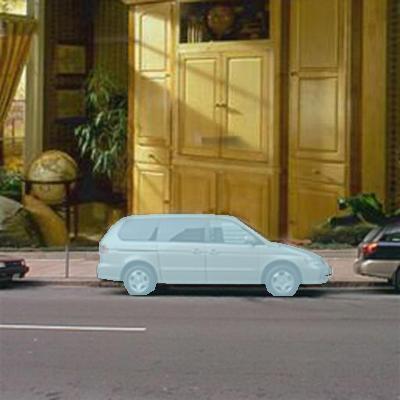} \\

        \includegraphics[width=0.155\linewidth, trim={30 20 50 20}, clip]{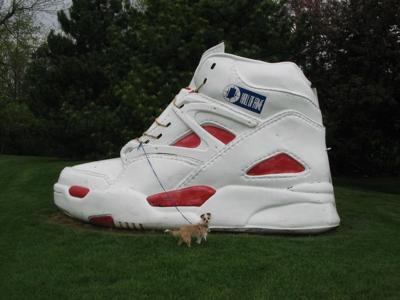} & 
        \includegraphics[width=0.155\linewidth, trim={30 20 50 20}, clip]{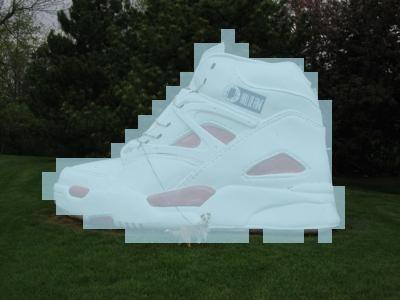} & 
        \includegraphics[width=0.155\linewidth, trim={30 20 50 20}, clip]{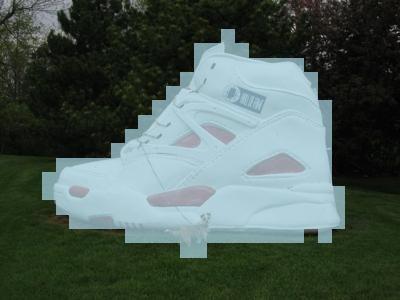} & 
        \includegraphics[width=0.155\linewidth, trim={30 20 50 20}, clip]{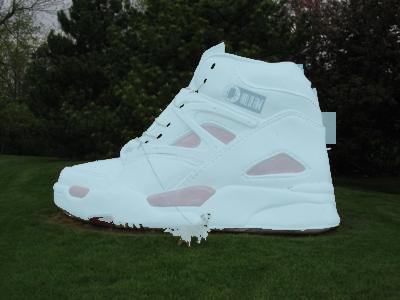} & 
        \includegraphics[width=0.155\linewidth, trim={30 20 50 20}, clip]{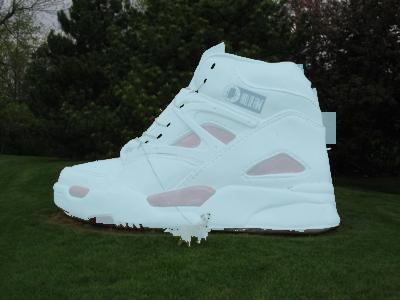} & 
        \includegraphics[width=0.155\linewidth, trim={30 20 50 20}, clip]{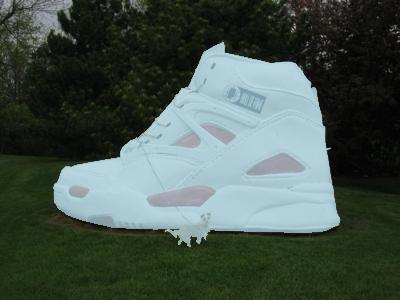} \\

        \includegraphics[width=0.155\linewidth, trim={20 10 20 120}, clip]{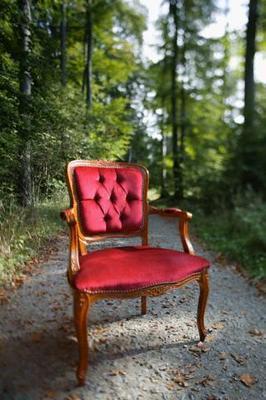} & 
        \includegraphics[width=0.155\linewidth, trim={20 10 20 120}, clip]{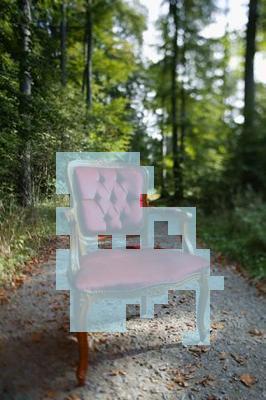} & 
        \includegraphics[width=0.155\linewidth, trim={20 10 20 120}, clip]{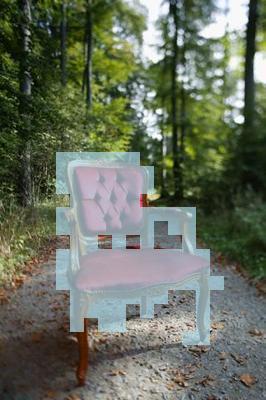} & 
        \includegraphics[width=0.155\linewidth, trim={20 10 20 120}, clip]{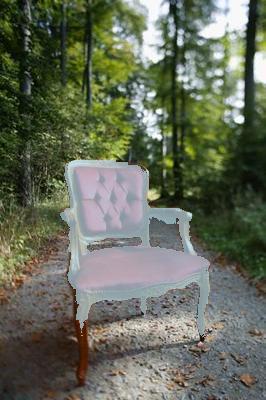} & 
        \includegraphics[width=0.155\linewidth, trim={20 10 20 120}, clip]{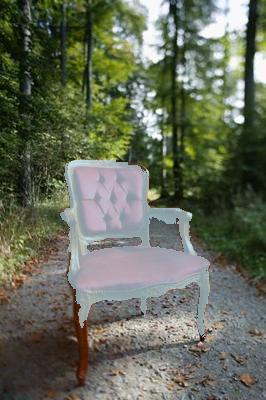} & 
        \includegraphics[width=0.155\linewidth, trim={20 10 20 120}, clip]{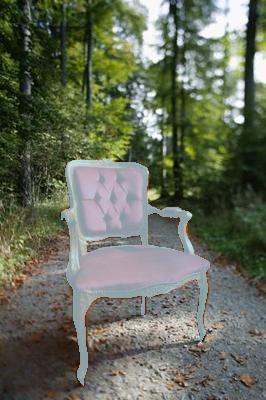} \\

        \includegraphics[width=0.155\linewidth, trim={30 0 90 60}, clip]{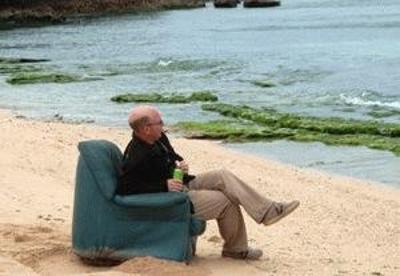} & 
        \includegraphics[width=0.155\linewidth, trim={30 0 90 60}, clip]{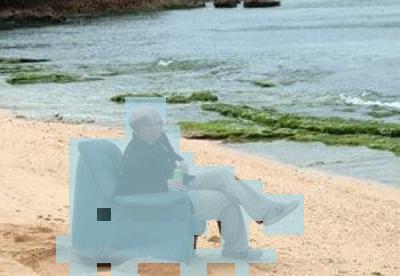} & 
        \includegraphics[width=0.155\linewidth, trim={30 0 90 60}, clip]{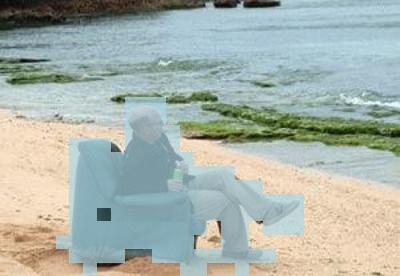} & 
        \includegraphics[width=0.155\linewidth, trim={30 0 90 60}, clip]{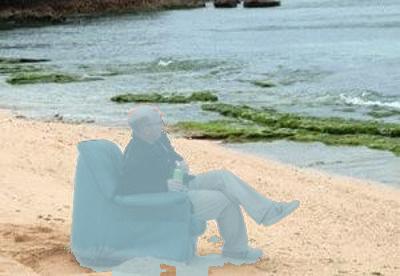} & 
        \includegraphics[width=0.155\linewidth, trim={30 0 90 60}, clip]{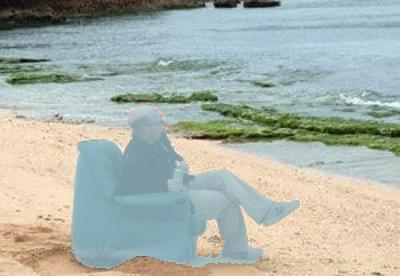} & 
        \includegraphics[width=0.155\linewidth, trim={30 0 90 60}, clip]{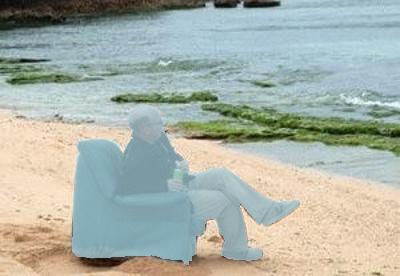} \\

        \includegraphics[width=0.155\linewidth, trim={0 0 100 100}, clip]{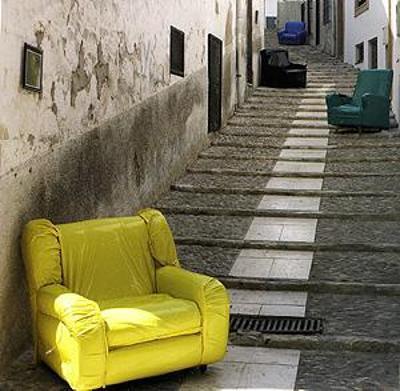} & 
        \includegraphics[width=0.155\linewidth, trim={0 0 100 100}, clip]{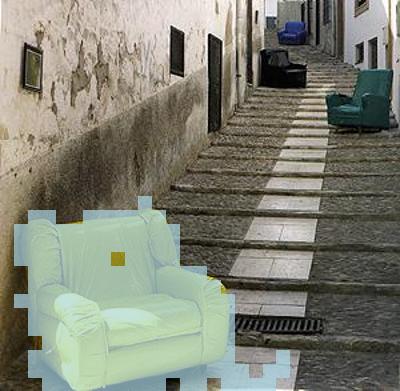} & 
        \includegraphics[width=0.155\linewidth, trim={0 0 100 100}, clip]{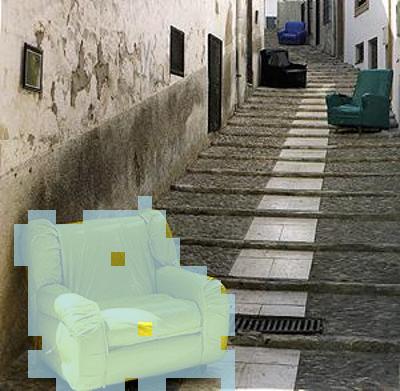} & 
        \includegraphics[width=0.155\linewidth, trim={0 0 100 100}, clip]{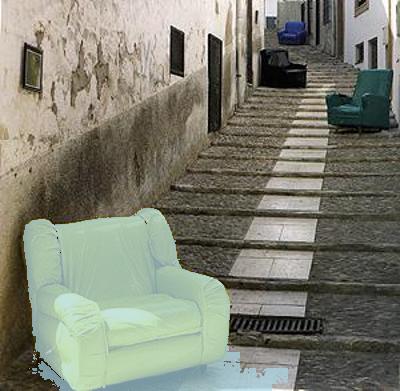} & 
        \includegraphics[width=0.155\linewidth, trim={0 0 100 100}, clip]{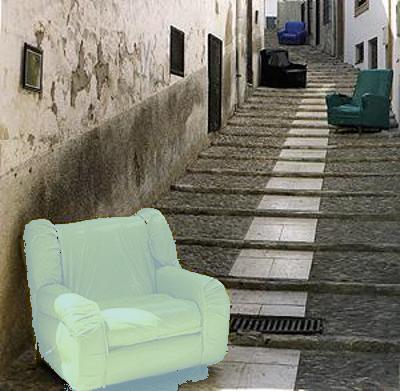} & 
        \includegraphics[width=0.155\linewidth, trim={0 0 100 100}, clip]{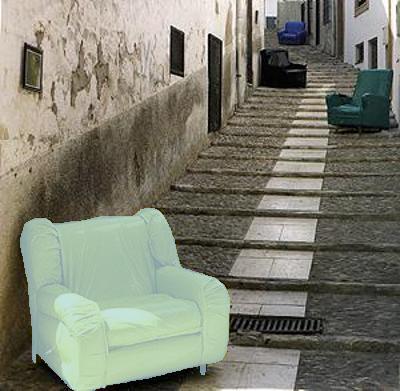} \\

    \end{tabular}
    \caption{Examples of salient segmentation from DUTOMRON~\citep{dutomron} using TokenCut~\citep{wang2023tokencut} with frozen CAPI~\citep{darcet2025cluster} on the raw embeddings (Raw pred.) and after correcting with \METHODNAME (\METHODNAME pred.). Suppressing positional noise with \METHODNAME either matches or improves the zero-shot saliency maps.
    We show the predictions per patch and after refining the segmentation maps with the bilateral solver (BS).  
    These examples are from the first samples in the dataset, and were not cherry picked except to show different outcomes of using \METHODNAME.
    }
    \label{fig:salient_improvement_examples_dutomron}
\end{figure*}

We show additional examples of salient segmentation using TokenCut~\citep{wang2023tokencut} in \cref{fig:salient_improvement_examples_duts,fig:salient_improvement_examples_dutomron}, for the DUTS~\citep{duts} and DUTOMRON~\citep{dutomron} datasets, respectively. The images displayed are from the first samples in the datasets, and were not cherry picked except to show different outcomes of using \METHODNAME in the case of DUTOMRON.
We show both the coarse per-patch prediction maps, and the refined maps after using the bilateral solver for edge aware post-processing; see \cref{sec:tokencut_details} for more TokenCut details.

